\documentclass[12pt]{article}
\usepackage{amsmath}
\usepackage{amsthm}
\usepackage{newtxmath}
\usepackage[T1]{fontenc}
\usepackage[utf8]{inputenc}
\usepackage{units}
\usepackage{graphicx}
\usepackage[authoryear]{natbib}
\usepackage{microtype}
\usepackage{zymacros}

\makeatletter

\usepackage{changepage}

 \usepackage[preprint]{nips_2018}

\usepackage{url}% simple URL typesetting
\usepackage{booktabs}% professional-quality tables
\usepackage{amsfonts}% blackboard math symbols
\usepackage{nicefrac}% compact symbols for 1/2, etc.
\usepackage{caption}
\usepackage{xcolor}

\author{
Wei Cai \\
\texttt{cai@smu.edu} \\
Dept of Mathematics, Southern Methodist University\\
Dallas, TX 75257, United States \\
\\
{\bf Zhi-Qin John Xu\thanks{Corresponding author}} \\
\texttt{xuzhiqin@sjtu.edu.cn} \\
School of Mathematical Sciences, \\MOE-LSC and Institute of Natural Sciences,\\ Shanghai Jiao Tong University,\\ Shanghai,      200240, P.R. China\\
}

\@ifundefined{showcaptionsetup}{}{%
 \PassOptionsToPackage{caption=false}{subfig}}
\usepackage{subfig}
\makeatother

\begin{document}

\title{Multi-scale Deep Neural Networks for Solving High Dimensional PDEs\thanks{October 24, 2019. }}

\maketitle

\begin{abstract}
In this paper, we propose the idea of radial scaling in frequency domain and activation functions with compact support to produce a multi-scale DNN (MscaleDNN), which will have the multi-scale capability in approximating high frequency and high dimensional functions and speeding up the solution of high dimensional PDEs. Numerical results on high dimensional function fitting and solutions of high dimensional PDEs, using loss functions with either Ritz energy or least squared PDE residuals,  have validated the increased power of multi-scale resolution and high frequency capturing  of the proposed MscaleDNN.
\end{abstract}
\section{Introduction}
Deep neural network (DNN) has found many application outside its traditional applications such as  image classification and speech recognition into the 
arena of scientific computing \citep{weinan2016deep, weinan2017deep,khoo2017solving,he2018relu,fan2018multiscale,han2017deep,xu2019frequency, huang2019int}, which may encounter data and solution in high dimensional spaces. The high dimensionality could come from the intrinsic physical models 
such as the Schrodinger equation for quantum many body systems or from the high dimensional stochastic variables due to the uncertainties in a physical systems such as those from random
media material properties and manufacturing stochastic variance. DNN has been recognized as a potential tool to handle the high dimensional problems from these applications.
However, to apply the commonly-used DNNs to these computational science and engineering problems, we are faced with several challenges. The most prominent issue is that the DNN normally only handles data with low frequency content well, it has been shown by a Frequency Principle (F-Principle) that many DNNs learn the low frequency content of the data quickly with good generalization error, but they
will be inadequate when high frequency data are involved \citep{xu_training_2018,xu2018understanding,xu2019frequency,rahaman2018spectral,zhang2019explicitizing,luo2019theory}. As a numerical approximation technique, this behavior of DNN is quite the opposite of that of the 
popular multigrid methods for solving PDEs where the convergence is achieved first in the high frequency spectrum of the solution. This feature of DNN will create problem 
when our physical phenomena demonstrate high frequency behavior such as high frequency wave
propagation in random media or excited states of a quantum system calculated with the Schrodinger equations of many particles. 

In this paper, we will propose a procedure to construct multi-scale DNNs, termed MscaleDNN, 
to speed up the approximation over a wide range of frequencies of
high dimensional functions and apply the resulting technique for the solution of
PDEs in high dimensions. The key idea is to find a way to convert the learning or approximation of high frequency data to that of a low frequency one. 
This approach has been attempted in a previous work in the development of a phase shift DNN (PhaseDNN) \citep{cai2019phasednn}, where the high frequency component of the data was given a phase shift downward to
a low frequency spectrum, the learning of the shifted data can be achieved with a small sized DNN quickly, which was then shifted (upward) to give approximation to the original high frequency data.
The PhaseDNN has been shown to be very effective to handle highly oscillatory data from solution of high frequency Helmholtz equation and functions of small dimensions. Due to
the phase shift employed along each coordinate direction, the PhaseDNN will have an intrinsic curse of dimensionality issue, therefore, can not be applied to high dimensional problems.
In this study, we will propose a different method to achieve the conversion of high frequency to lower one by using a radial partition of the Fourier space and applying a scaling down
operation to low frequency to learn high dimensional functions. because of the scaling operation along the radial direction in the k-space, this approach naturally avoids the curse of dimensionality issue of the PhaseDNN.

The rest of the paper is organized as follows. In section 2, we will introduce the idea of frequency scaling to generate a multi-scale DNN representation as well as that of compact supported activation function, the latter will allow the multi-scale resolution capability of the resulting DNNs. Section 3 will present two minimization approaches for finding the solution of elliptic PDEs: one through the Ritz energy,  the other through least square residual error of the PDEs. Next, numerical results of high dimensional and high frequency function fitting as well as the solution of high dimensional PDEs by the proposed MscaleDNN will be given in Section 4. Finally, Section 5 gives a conclusion and some discussion for further work.

\section{The idea of frequency scaled DNN and activation function with compact support}

Our overall goal is to find the solution to the following high dimensional elliptic PDE,%
\begin{equation}
-\epsilon\nabla^{2}u+V(r)u=f(\mathbf{r}),\text{ \ }\mathbf{r}\in\Omega\subset
R^{d},d>>1,\nonumber
\end{equation}
where $\epsilon=\frac{h^{2}}{2m}$ for Schrodinger equations, $V(\mathbf{r})$
from the external potential such as from the nucleus and external excitations.

As the solution $u(\mathbf{r})$ can be very high dimensional and we hope to find a
DNN approximation to
\begin{equation}
T^{\ast}(\mathbf{r},\mathbf{\theta})\sim u(\mathbf{r}).\label{PDNN}%
\end{equation}

In this section, we will discuss a naive idea how to use phase
scaling in Fourier wave number space to reduce a high frequency learning
problems to a low frequency learning for the DNN and will also point out the
difficulties it may encounter as a practical algorithm.

Consider a band-limited function $f(\mathbf{r}),\mathbf{r\in}R^{d}$ whose
Fourier transform $\widehat{f}(\mathbf{k})$ has a compact support, i.e.,%

\begin{equation}
\sup\widehat{f}(\mathbf{k})\subset B(K_{\text{max}})=\{\mathbf{k\in}%
R^{d},|\mathbf{k|\leq}K_{\text{max}}\}.\label{support}%
\end{equation}

We will first partition the domain $B(K_{\text{max}})$ as union of $M$
concentric annulus with uniform or non-uniform width, e.g.,

\begin{equation}
A_{i}=\{\mathbf{k\in}R^{d},(i-1)K_{0}\leq|\mathbf{k|\leq}iK_{0}\},K_{0}%
=K_{\text{max}}/M,1\leq i\leq M\label{anulnus}%
\end{equation}
so that %

\begin{equation}
B(K_{\text{max}})=%
%TCIMACRO{\dbigcup \limits_{i=1}^{M}}%
%BeginExpansion
{\displaystyle\bigcup\limits_{i=1}^{M}}
%EndExpansion
A_{i}.\label{part}%
\end{equation}

Now, we can decompose the function $\widehat{f}(\mathbf{k})$ as follows%

\begin{equation}
\widehat{f}(\mathbf{k})=%
%TCIMACRO{\dsum \limits_{i=1}^{M}}%
%BeginExpansion
{\displaystyle\sum\limits_{i=1}^{M}}
%EndExpansion
\chi_{A_{i}}(\mathbf{k})\widehat{f}(\mathbf{k})\triangleq%
%TCIMACRO{\dsum \limits_{i=1}^{M}}%
%BeginExpansion
{\displaystyle\sum\limits_{i=1}^{M}}
%EndExpansion
\widehat{f}_{i}(\mathbf{k}),\label{POU}%
\end{equation}
where
\begin{equation}
\sup\widehat{f}_{i}(\mathbf{k})\subset A_{i}.\label{SupAi}%
\end{equation}

\bigskip The decomposition in the $k$-space give a corresponding one in the
physical space%

\begin{equation}
f(\mathbf{r})=%
%TCIMACRO{\dsum \limits_{i=1}^{M}}%
%BeginExpansion
{\displaystyle\sum\limits_{i=1}^{M}}
%EndExpansion
f_{i}(\mathbf{r}),\label{Partx}%
\end{equation}
where %

\begin{equation}
f_{i}(\mathbf{r})=\mathcal{F}^{-1}[\widehat{f}_{i}(\mathbf{k})](\mathbf{r}%
)=f(\mathbf{r})\ast\chi_{A_{i}}^{\vee}(\mathbf{r}),\label{convol}%
\end{equation}
and the inverse Fourier transform of $\chi_{A_{i}}(\mathbf{k})$ is called the
frequency selection kernel \citep{cai2019phasednn} and can be computed analytically using
Bessel functions%

\begin{equation}
\chi_{A_{i}}^{\vee}(\mathbf{r})=\frac{1}{(2\pi)^{d/2}}%
%TCIMACRO{\dint \limits_{A_{i}}}%
%BeginExpansion
{\displaystyle\int\limits_{A_{i}}}
%EndExpansion
e^{i\mathbf{k\circ r}}d\mathbf{r.}\label{kernel}%
\end{equation}

From (\ref{SupAi}), we can apply a simple down-scaling to convert the high
frequency region $A_{i}$ to a low frequency region. Namely, we define a scaled
version of $\widehat{f}_{i}(\mathbf{k})$ as%

\begin{equation}
\widehat{f}_{i}^{(\text{scale})}(\mathbf{k})=\widehat{f}_{i}(\alpha
_{i}\mathbf{k}),\alpha_{i}>1,\label{fkscale}%
\end{equation} 

\bigskip and%

\begin{equation}
f_{i}^{(\text{scale})}(\mathbf{r})=f_{i}(\frac{1}{\alpha_{i}}\mathbf{r}%
),\label{fscale}%
\end{equation}
or 
\begin{equation}
f_{i}(\mathbf{r})=f_{i}^{(\text{scale})}(\alpha_{i}\mathbf{r}),
\label{fscale_inv}%
\end{equation}
noting the low frequency spectrum of the scaled function if $\alpha_i$ is chosen large enough, i.e.,
\begin{equation}
\sup\widehat{f}_{i}^{(\text{scale})}(\mathbf{k})\subset \{\mathbf{k\in}R^{d},\frac{(i-1)K_{0}}{\alpha_{i}}\leq|\mathbf{k|\leq}\frac{iK_{0}}{\alpha_{i}}\}.\label{fssup}%
\end{equation}

Using the F-Principle of common DNNs \citep{xu2019frequency}, with $iK_{0}/\alpha_{i}$ being small, we can train a DNN $h_{i}%
(\mathbf{r,\theta}^{n_{i}})$ to learn  $f_{i}^{(\text{scale})}(\mathbf{r})$ quickly

\begin{equation}
f_{i}^{(\text{scale})}(\mathbf{r})\sim h_{i}(\mathbf{r,\theta}^{n_{i}%
}),\label{DNNi}%
\end{equation}
which gives an approximation to $f_{i}(\mathbf{r})$ immediately%

\begin{equation}
f_{i}(\mathbf{r})\sim h_{i}(\alpha_{i}\mathbf{r,\theta}^{n_{i}})\label{fi_app}%
\end{equation}
and to $f(\mathbf{r})$ as well %

\begin{equation}
f(\mathbf{r})\sim%
%TCIMACRO{\dsum \limits_{i=1}^{M}}%
%BeginExpansion
{\displaystyle\sum\limits_{i=1}^{M}}
%EndExpansion
h_{i}(\alpha_{i}\mathbf{r,\theta}^{n_{i}}).\label{f_app}%
\end{equation}

\bigskip The difficulty of the above procedure for approximating function in
high dimension is the need to compute the
$d$-dimensional convolution in (\ref{convol}), which leads to the issue of the curse of
dimensionality.

\noindent{\bf Compact supported activation function}. In order to produce scale separation and identification capability of the MscaleDNN, we take
the hint from the theory of compact mother scaling function in the wavelet theory \citep{daubechies1992ten} and consider 
the activation functions with a compact support. This way if the activation function is scaled by a factor $\alpha$, the scaled activation function will have frequency spectrum as $\alpha$ times that of that of the original activation function, allowing the corresponding neuron in the MscaleDNN to learn more easily the corresponding frequency of the solution.

 The support of the common activation function $\ReLU (x)=\max (0,x)$ is not compact.  To produce an activation function with compact support, we simply use the following modified activation function. 
\[
{\rm sReLU}(x)={\rm ReLU}(-(x-1))\times{\rm ReLU}(x).
\]
The support of $\sReLU$ is on $[0,1]$. In order to have a differentiable activation function, we can use ${(\rm sReLU}(x))^2$  and ${(\rm sReLU}(x))^3$ to have continuous first and second derivatives, respectively, which are shown in  Fig.~\ref{fig:relu}.

\begin{center}
\begin{figure}[h]
\begin{centering}
	\subfloat[$\ReLU(x)$]{\begin{centering}
				\includegraphics[scale=0.35]{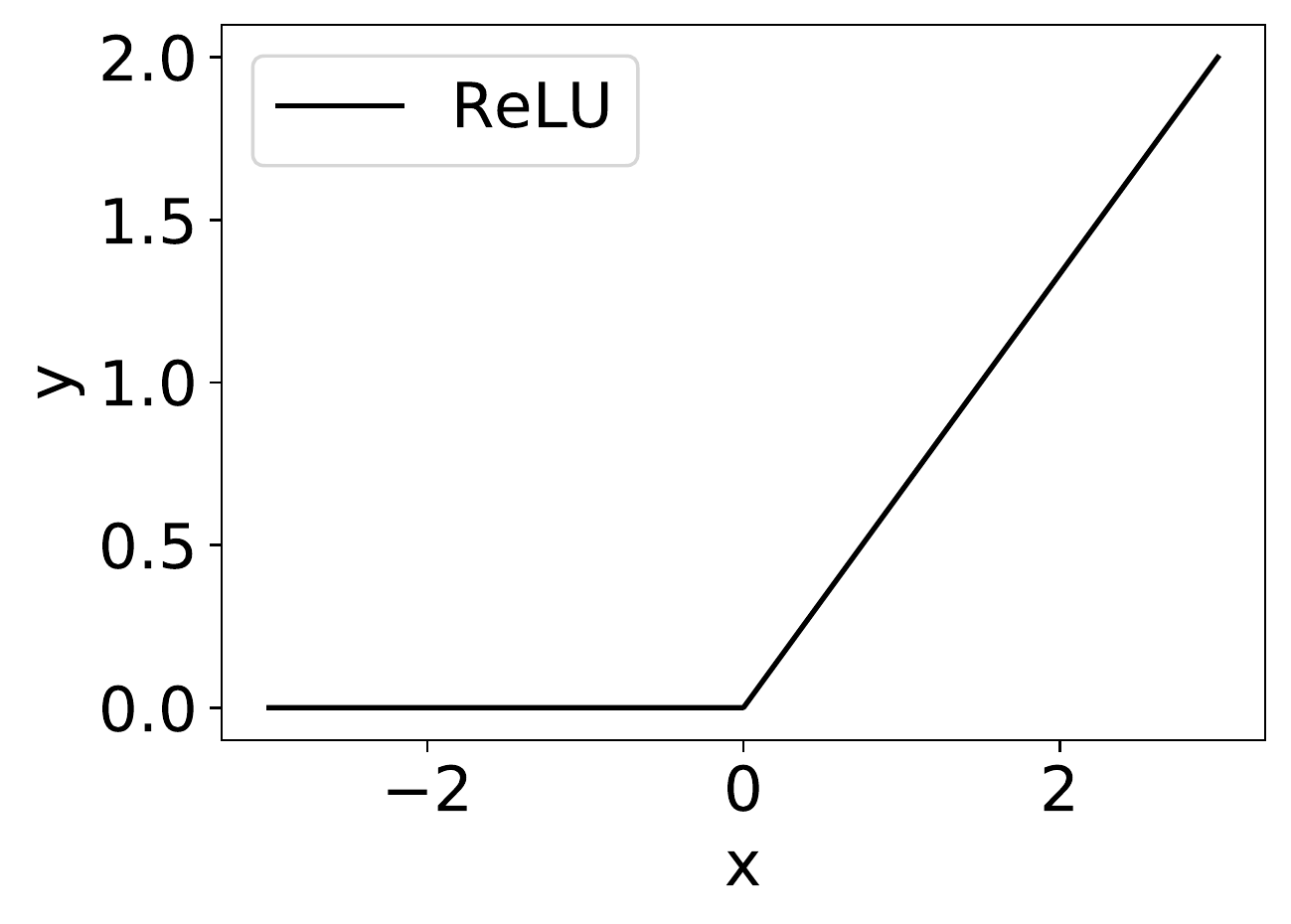}
	\par\end{centering}}
	\subfloat[$\sReLU(x)$]{\begin{centering}
				\includegraphics[scale=0.35]{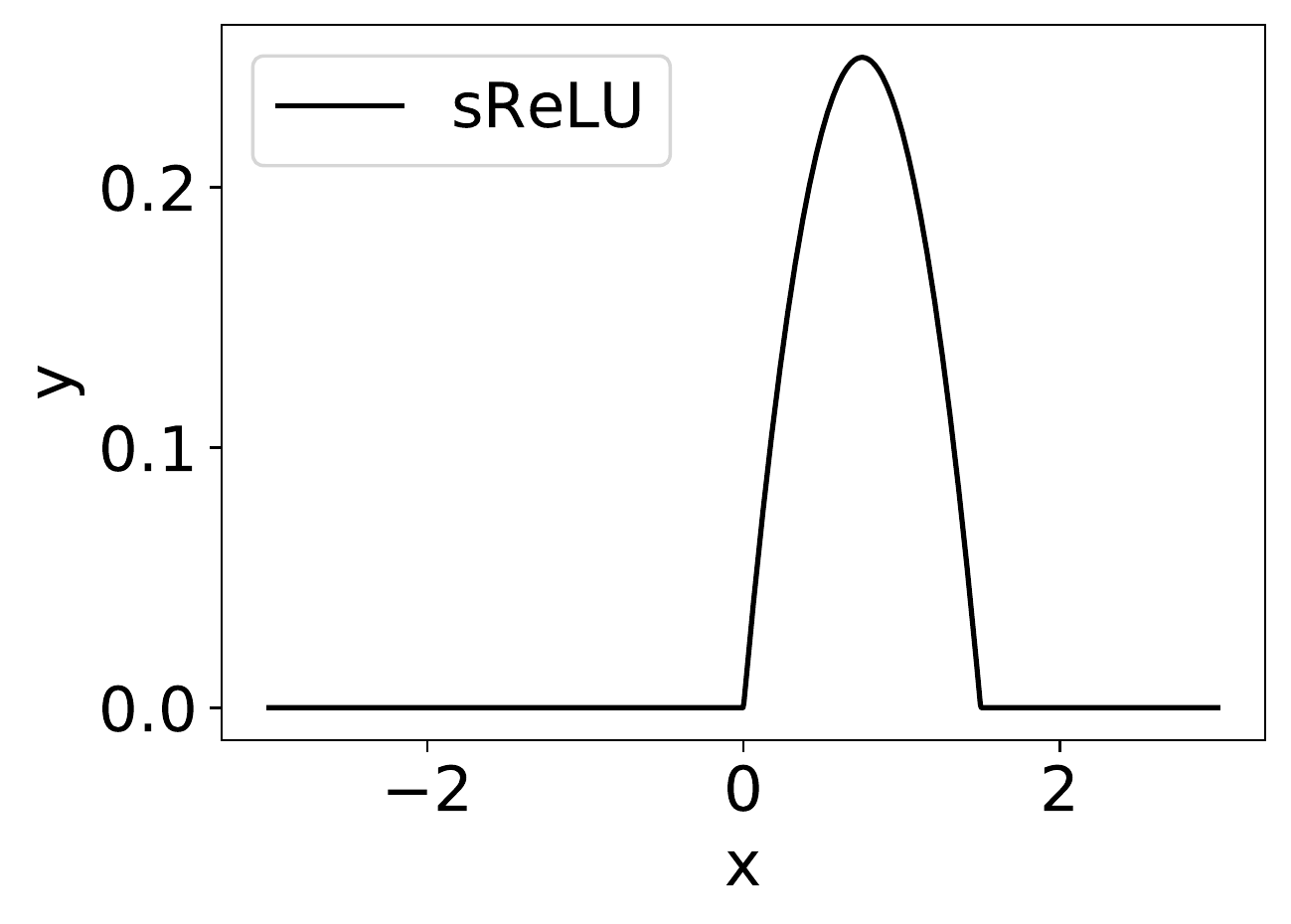}
	\par\end{centering}}
	
	\subfloat[$(\sReLU(x))^2$]{\begin{centering}
				\includegraphics[scale=0.35]{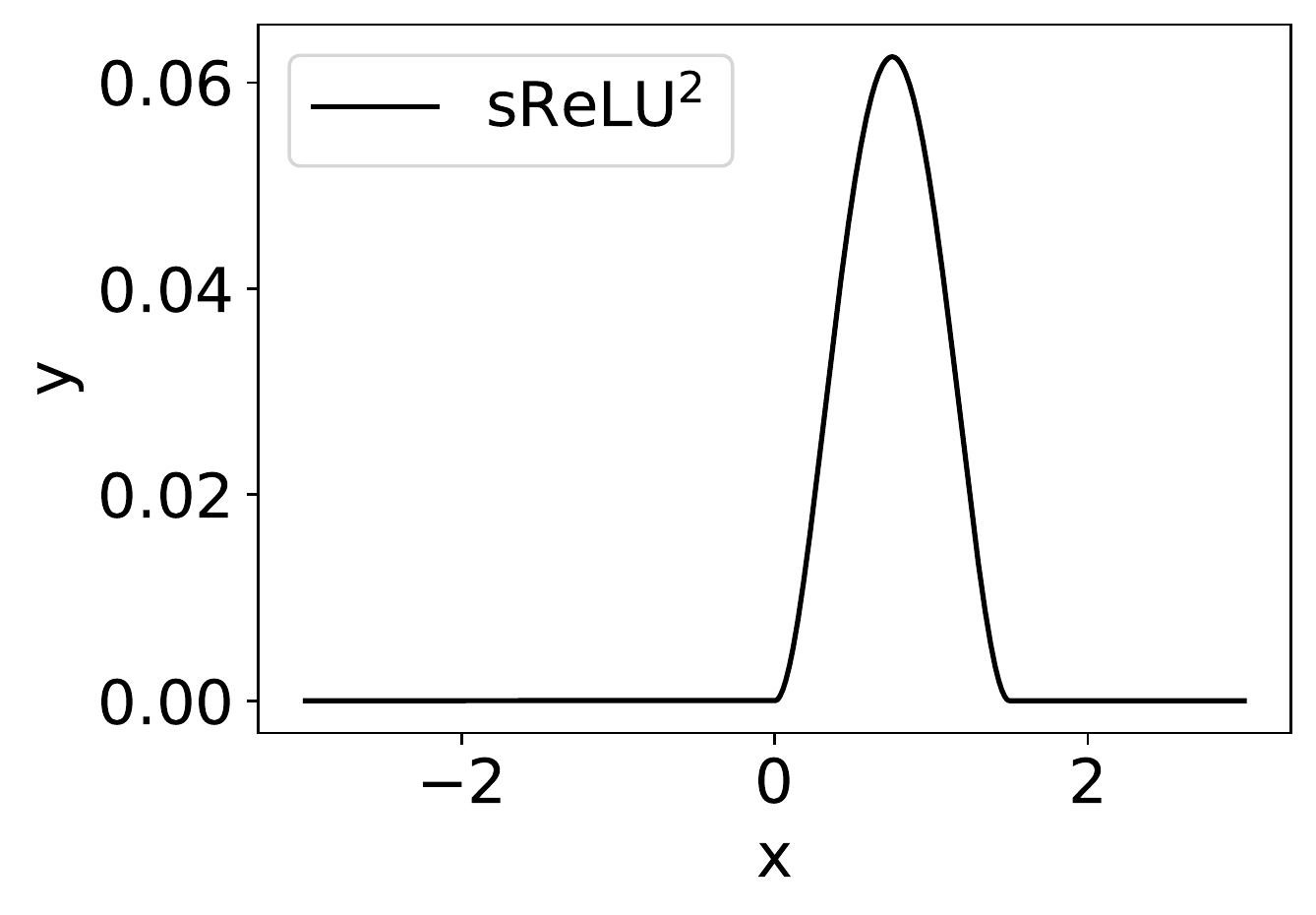}
	\par\end{centering}}
	\subfloat[$(\sReLU(x))^3$]{\begin{centering}
				\includegraphics[scale=0.35]{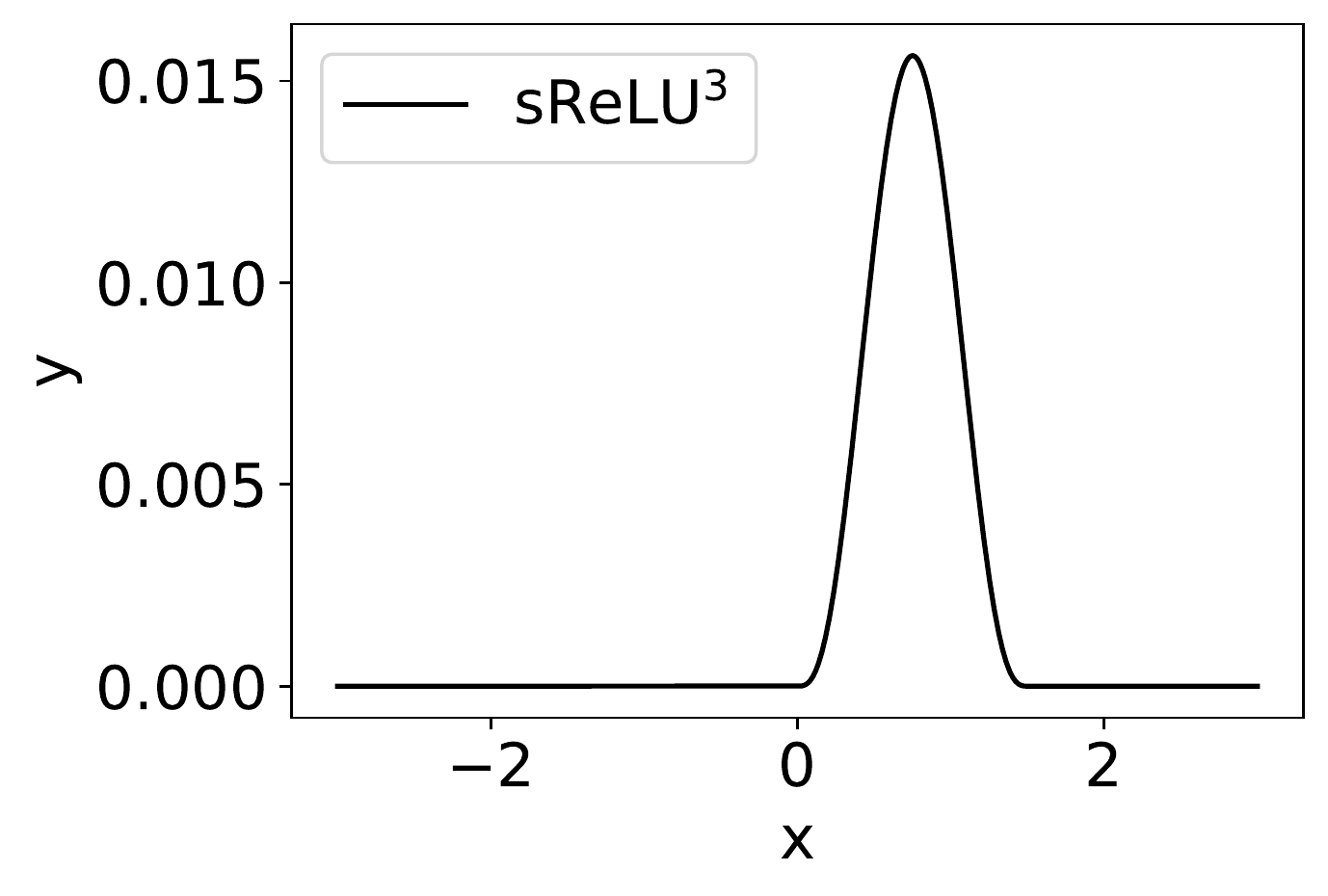}
	\par\end{centering}}

\par\end{centering}
\caption{ \label{fig:relu}}
\end{figure}
\par\end{center}

The weights in the MscaleDNN neurons are initialized by a normal 
distribution, namely, the weights  are sampled from 
${\cal D}_{1}=\mathcal{N}(0,(2/(n_{{\rm in}}+n_{{\rm out}}))^2)$ or ${\cal D}_{2}=\mathcal{N}(0,2/(n_{{\rm in}}+n_{{\rm out}}))$, where $n_{{\rm in}}$
and $n_{{\rm out}}$ are the input and output dimensions of the corresponding 
layer, respectively. Note that these two initialization methods are widely used and studied  \citep{glorot2010understanding,jacot2018neural,rotskoff2018parameters}. The DNNs are trained by Adam \citep{kingma2014adam}.

\section{MscaleDNN with compact supported activation function}

Even though the procedure leading to (\ref{f_app}) is not practical
for high dimensional approximation, however it does suggest a plausible
form of function space for finding the solution more quickly with DNN functions. We can use a series of $\alpha_{i}$ ranging from $1$ to a large number to produce a MscaleDNN, which can achieve our goal to
speed up the convergence for solution with many frequencies with better accuracy.

\noindent{\bf MscaleDNN structure}.  To realize the MscaleDNN, we separate the neuron in the first hidden-layer into to $A$ parts. The  neuron in the $i$-th part receives input $i\vx$, that is,  its output is $\sigma(i\vw\cdot\vx+b)$, where $\vw$, $\vx$, b are weight, input and bias term, respectively. The complete MscaleDNNs  reads as
\begin{equation}
    h(\vx) = \vW_L \sigma\circ(\mW_{L-1}\sigma\circ(\cdots \sigma\circ(\mW_{1}\sigma\circ(\vK\odot\mW_0 \vx+\vb_0)+\vb_1)\cdots)+\vb_{L-1})
\end{equation}
where $\vx\in\mathbb{R}^d$, $\mW_l\in\mathbb{R}^{n_{l+1}\times n_{l}}$, $n_l$ is the neuron number of $l$-th hidden layer, $n_0=d$, $b_l\in\mathbb{R}^{l+1}$,
$\sigma$ is a scalar function and ``$\circ$'' means entry-wise operation, $\odot$ is the Hadamard product and 
\begin{equation}
\vK=(\underbrace{1,1,\cdots,1}_{\text{1st part}},2,\cdots,i-1,\underbrace{i,i,\cdots,i}_{\text{ith part}},i+1,\cdots,\underbrace{A,A,\cdots,A}_{\text{Ath part}})^T.
\end{equation} 
Fig.~\ref{fig:msdnnex} shows a  toy example of $A=3$. The only difference compared with a normal fully-connected network is the input to the first hidden layer.

\begin{center}
\begin{figure}[h]
\begin{centering}
	\includegraphics[scale=0.4]{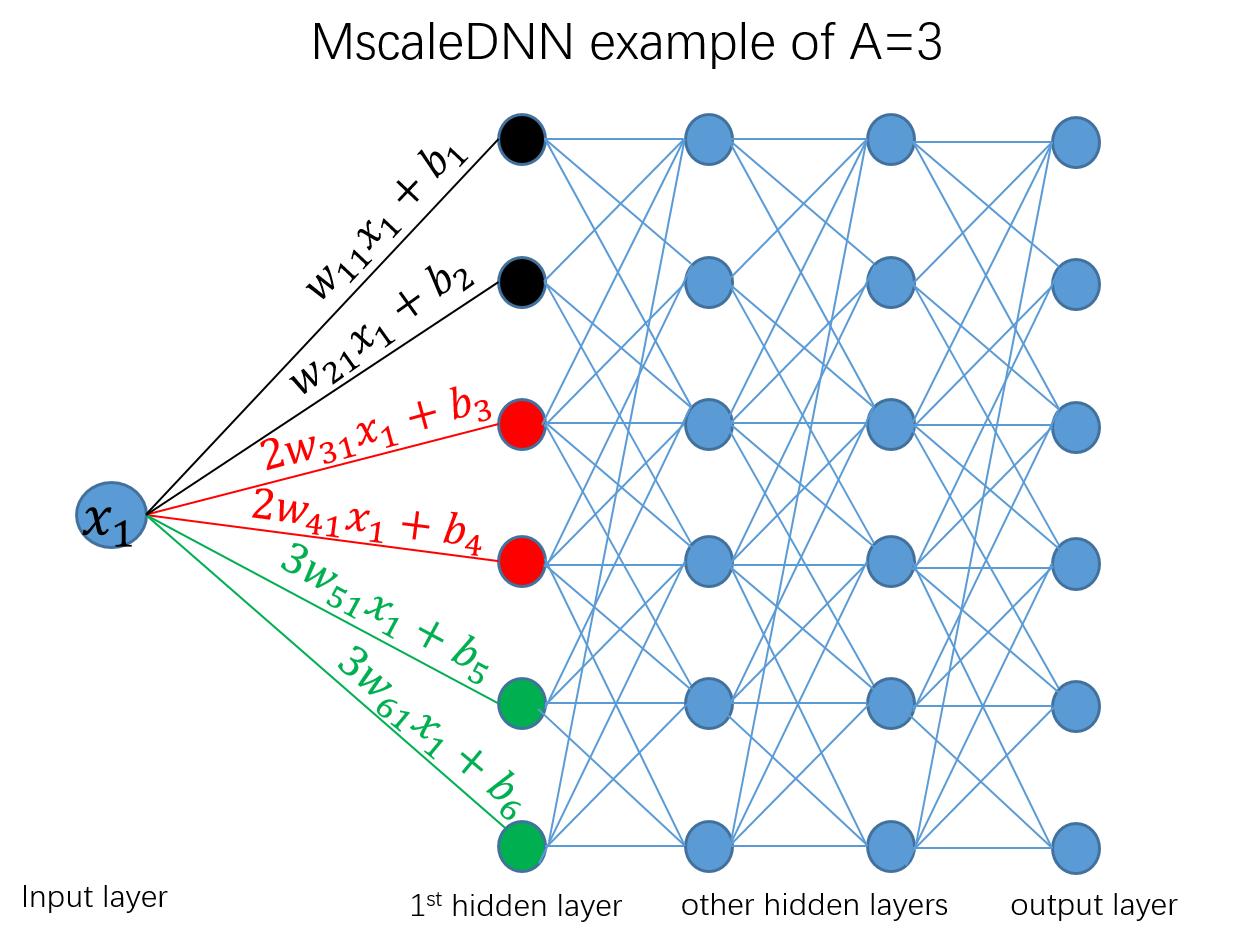}
\par\end{centering}
\caption{Illustration of a MscaleDNN. \label{fig:msdnnex}}
\end{figure}
\par\end{center}

We would use MscaleDNN to fit high-dimensional functions and high-frequency functions. We will also introduce two loss functions of using MscaleDNN for finding the solution of  PDEs. We consider the following problem,
\begin{equation}
-\Delta u(\vx)=g(\vx),\quad \vx \in \Omega,
\end{equation}

with the  boundary condition 
\begin{equation}
u(\vx)=\tilde{g}(\vx),\quad \vx \in \partial \Omega.
\end{equation}
where $\vx=(x_1,x_2,\cdots,x_d)$. We denote the true solution as $u_{\rm true}(\vx)$. In each training step, we sample $n$ points in $\Omega$   and $\tilde{n}$ points at each hyperplane that composes  $\partial \Omega$.
The first loss function is based on the variational Ritz formulation of the elliptic problem  and the second one is based on the mean square of the residual of the differential equation.

\subsection{ A Ritz variational method for PDE}

 The deep Ritz method as proposed by  \cite{weinan2018deep} produces a variational solution $u(\mathbf{r})$ through the following minimization problem%

\begin{equation}
u=\arg\min_{v\in\Pi}J(v),\label{Ritz}%
\end{equation}
where the energy functional is defined as%

\begin{equation}
J(v) =\int_{\Omega}
\left(  \frac{1}{2}\epsilon(\mathbf{r})|\nabla v|^{2}+V(r)v^{2}\right)  dr-
{\displaystyle\int\limits_{\Omega}}
%EndExpansion
f(\mathbf{r})v(\mathbf{r})dr
 \triangleq%
%TCIMACRO{\dint \limits_{\Omega}}%
%BeginExpansion
{\displaystyle\int\limits_{\Omega}}
%EndExpansion
E(\mathbf{v(r))dr}.\label{energy}
\end{equation}

We use MscaleDNN output $h(x,\theta)$ to parameterize $v$ in the above problem, where $\theta$ is the DNN parameter set. Then, the MscaleDNN solution is 

\begin{equation}
\mathbf{\theta}_{\ast}=\arg\min_{\theta
}J(h(x,\theta)),\label{Ritz}%
\end{equation}

The minimizer $\mathbf{\theta}_{\ast}$ can be found by stochastic gradient decent (SGD) as in \cite{weinan2018deep},%

\begin{equation}
\mathbf{\theta}^{(n+1)}=\mathbf{\theta}^{(n)}+\eta\nabla_{\mathbf{\theta}%
}J(h(x,\theta)). \label{sgd}%
\end{equation} 
The integral in Eq. (\ref{energy}) can not be integrated due to high dimensional curse, and in fact, will only be sampled at some random points at each training step  (see (2.11) in \cite{weinan2018deep}), that is,
\begin{align}
\nabla_{\mathbf{\theta}}J(h(x,\theta))\sim\nabla_{\mathbf{\theta}}\frac{1}{Ns}%
%TCIMACRO{\dsum \limits_{s=1}^{N_{s}}}%
%BeginExpansion
{\displaystyle\sum\limits_{s=1}^{N_{s}}}
%EndExpansion
E(h(\mathbf{r}_{s},\theta)).\label{GradJ}
\end{align}

At convergence $\mathbf{\theta}^{(n)}\rightarrow\mathbf{\theta}_{\ast}$, we
obtain a MscaleDNN solution $h(x,\theta_{\ast})$ for the given material constant $\mathbf{\epsilon
}^{(s)}$.

In our numerical test of this paper, the Ritz loss function is chosen to be
\begin{equation}
L_{\rm ritz}(h)=\frac{1}{n}\sum_{\vx\in S}(|\nabla h(\vx)|^2/2-g(\vx)h(\vx))+\beta*\frac{1}{\tilde{n}}\sum_{\vx\in \tilde{S}}(h(\vx)-\tilde{g}(\vx))^2. \label{ritzlossnum}
\end{equation}

where $h(\vx)$ is the DNN ouput, $S$ is the sample set from $\Omega$ and   $n$ is the sample size. $\tilde{n}$  indicates sample set from $\partial\Omega$. The second penalty term  is to enforce the boundary condition.

\subsection{MscaleDNN least square error method for PDEs}

In an alternative approach, we can simply use the loss function of Least Squared Residual Error (LSE) ,
\begin{equation}
L_{\rm LSE}(h)=\frac{1}{n}\sum_{\vx\in S}(\Delta h(\vx)+g(\vx))^2+\beta*\frac{1}{\tilde{n}}\sum_{\vx\in \tilde{S}}(h(\vx)-\tilde{g}(\vx))^2.\label{lselossnum}
\end{equation}

To see the learning accuracy, we also compute the distance between $h(\vx)$ and $u_{\rm true}$,
\begin{equation}
{\rm MSE}(h(\vx),u_{\rm true}(\vx))=\frac{1}{n+\tilde{n}}\sum_{\vx\in S \cup \tilde{S}}(h(\vx)-u_{\rm true}(\vx))^2
\end{equation}

\section{Numerical experiments}

 In the following, we use solid line to indicate the train loss and dashed line to indicate the test loss. The curve of the training loss often overlaps well with that of the test loss for most cases, indicating that the neural networks generalize well.

We explain the legends in figures here. ``ms1'', ``ms1''  and  ``ms100'' indicate the scale number $A$ of $1$, $10$ and $100$, respectively. ``ReLU'' and ``sReLU'' indicate the activation functions. ``train'' and ``test'' indicate the training loss and the test loss, respectively. All networks are trained by Adam \citep{kingma2014adam}.

\subsection{Fitting 3-dimensional functions}
In this section, we use a 3d function to show that MscaleDNN can learn target function much faster than the normal one-scale network. In addition, we also show the DNN with compact-supported activation function learns faster than that of normal non-compact-supported  activation function. 
We use DNNs to fit a 3d oscillatory function: $\mathbb{R}^{3}\rightarrow\mathbb{R}$, that is,
\begin{equation}
f(\boldsymbol{x})=\sum_{j=1}^{3}\cos(10x_{j})+\sin(5x_{j}),\quad x_{j}\in[-\pi/2,\pi/2],\label{eq:sinx}
\end{equation}
where $\boldsymbol{x}=(x_{1},x_{2},x_{3})$. 

From the experiment of one-hidden layer network initialized by ${\cal D}_{1}$ in Fig.~\ref{fig:fithd_srelu} and the multiple hidden-layer network initialized by ${\cal D}_{2}$ in Fig.~\ref{fig:fithd_srelu2}, we can observe the following behaviors. First, the DNN with activation function $\sReLU$ learns much faster than that with $\ReLU$. Second, $\sReLU$-DNN with $100$ scales is much faster than that with only $1$ scale.

\begin{center}
\begin{figure}[h]
\begin{centering}
	\includegraphics[scale=0.5]{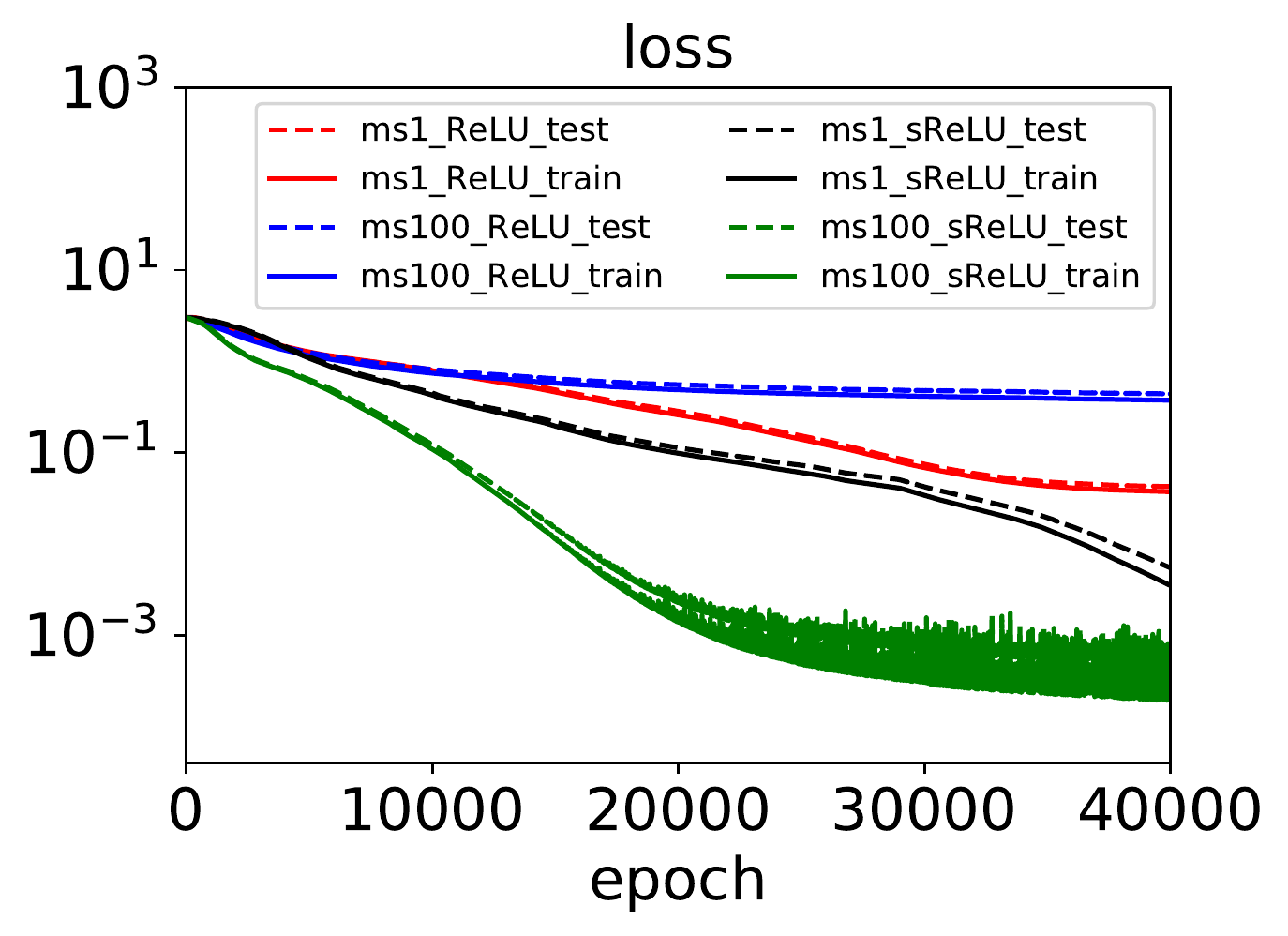}
\par\end{centering}
\caption{Loss function vs. training epoch.  We use a network 3-2500-1 with activation function $\ReLU(x)$ or  $\sReLU(x)$ indicated by the legend. The learning rate is  $5\times10^{-5}$ with a decay rate  $2\times10^{-7}$ for each full-batch training step. The training and test dataset are both $10000$ random samples. Weights are initialized by  ${\cal D}_{1}$. \label{fig:fithd_srelu}}
\end{figure}
\par\end{center}

\begin{center}
\begin{figure}[h]
\begin{centering}
	\includegraphics[scale=0.5]{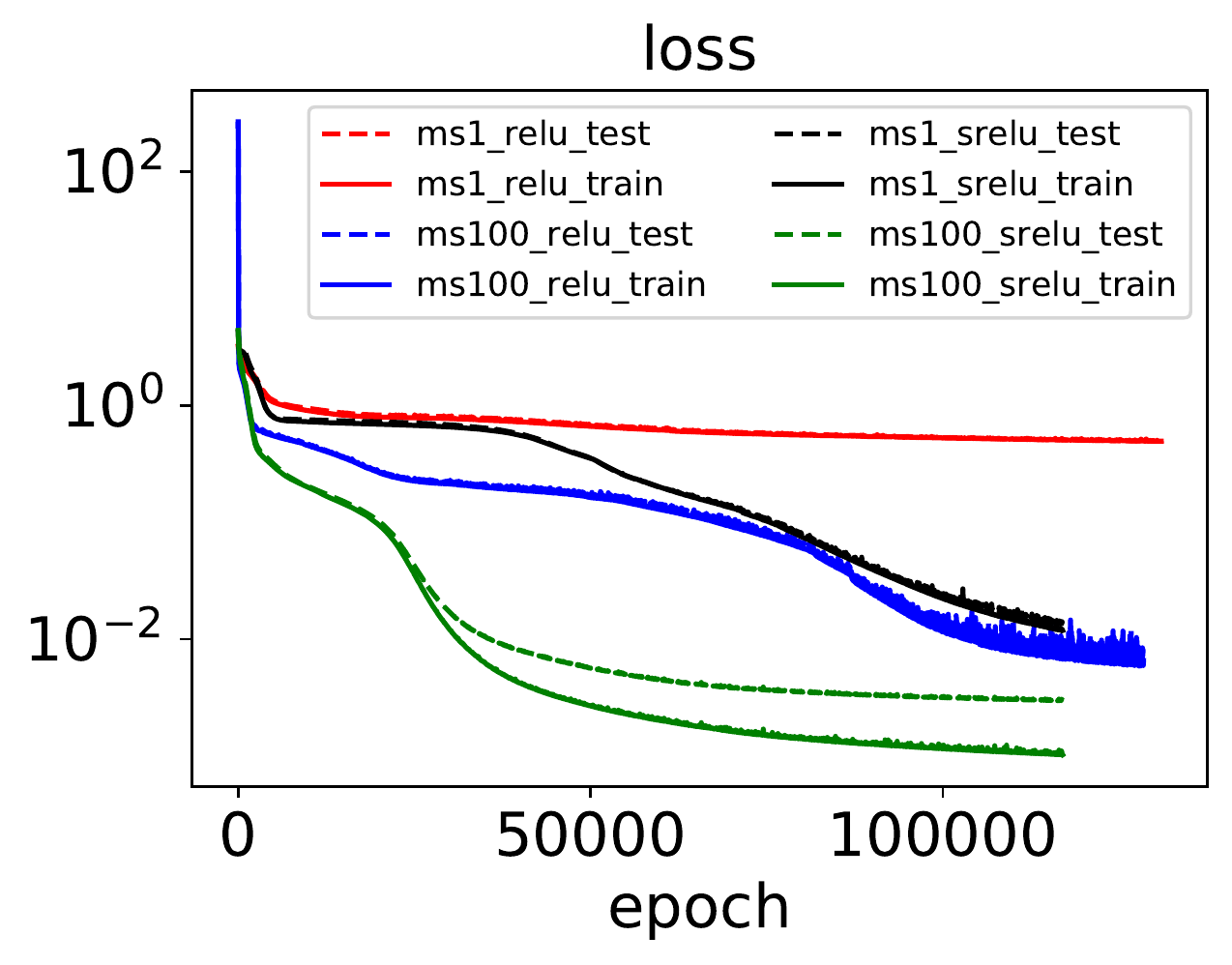}
\par\end{centering}
\caption{Loss function vs. training epoch.  We use a network 3-500-500-500-500-1 with activation function $\ReLU(x)$ or  $\sReLU(x)$ indicated by the legend. The learning rate is  $3\times10^{-6}$ with a decay rate  $5\times10^{-7}$ for each  training step with batch size 1000. The training and test dataset are both $5000$ random samples. Weights are initialized by  ${\cal D}_{2}$. \label{fig:fithd_srelu2}}
\end{figure}
\par\end{center}

\subsection{Fitting 60-dimensional functions with intrinsic low dimensional structure}

Real data are often high-dimensional, however, its intrinsic dimension is often low. In this subsection, we use DNNs to fit 60-dimensional data, which has a intrinsic dimension 3.  Consider 
\begin{equation}
f(\vx(\vt))=\sum_{j=1}^{d_{in}}\cos(10t_{j})+\sin(5t_{j}),\quad t_{j}\in[0,1],\label{eq:highloweq}
\end{equation}
where $\vx=(x_{1},x_{2},\cdots,x_{d})$,  $\vx=(x_{1},x_{2},\cdots,x_{d})$,  $d_{in}=3$ and $d=60$. 
We use a DNN of 
d-200-200-200-1. We would consider a linear embedding and a non-linear embedding situation.

For the linear embedding example, we consider 
\begin{equation}
x_{i}=\cos(i)t_{[i/(d/d_{in})]},
\end{equation}
where $[\cdot]$ is the operation of taking the integer part. 

For the non-linear embedding example, we consider, 
\begin{equation}
x_{i}=\cos(\cos(i)t_{[i/(d/d_{in})]}).
\end{equation}
For the networks initialized by ${\cal D}_{1}$, as shown by the training loss functions in Fig. \ref{fig:DNN-highd} (a, b), the MscaleDNN with sReLU is faster in learning both  linear and nonlinear embedded data. Note that the MscaleDNN with ReLU does not show advantage with more scales, as shown in Fig. \ref{fig:DNN-highd} (c). For the networks initialized by ${\cal D}_{2}$, as shown in Fig. \ref{fig:DNN-highd2}, results are similar.

\begin{center}
\begin{figure}[h]
\begin{centering}
	\subfloat[$\sReLU(x)$, linear]{\begin{centering}
				\includegraphics[scale=0.35]{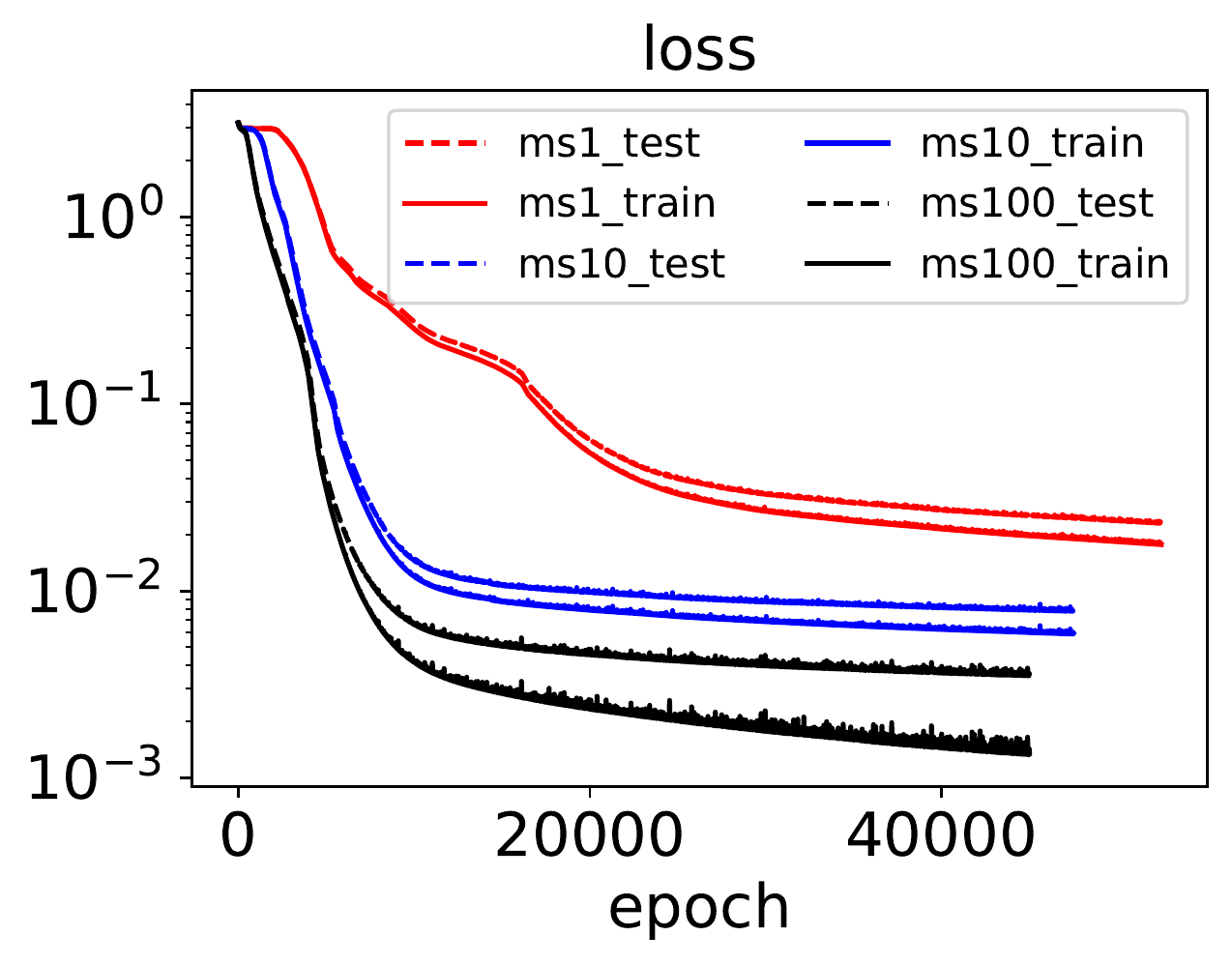}
	\par\end{centering}}
	\subfloat[$\sReLU(x)$, non-linear]{\begin{centering}
				\includegraphics[scale=0.35]{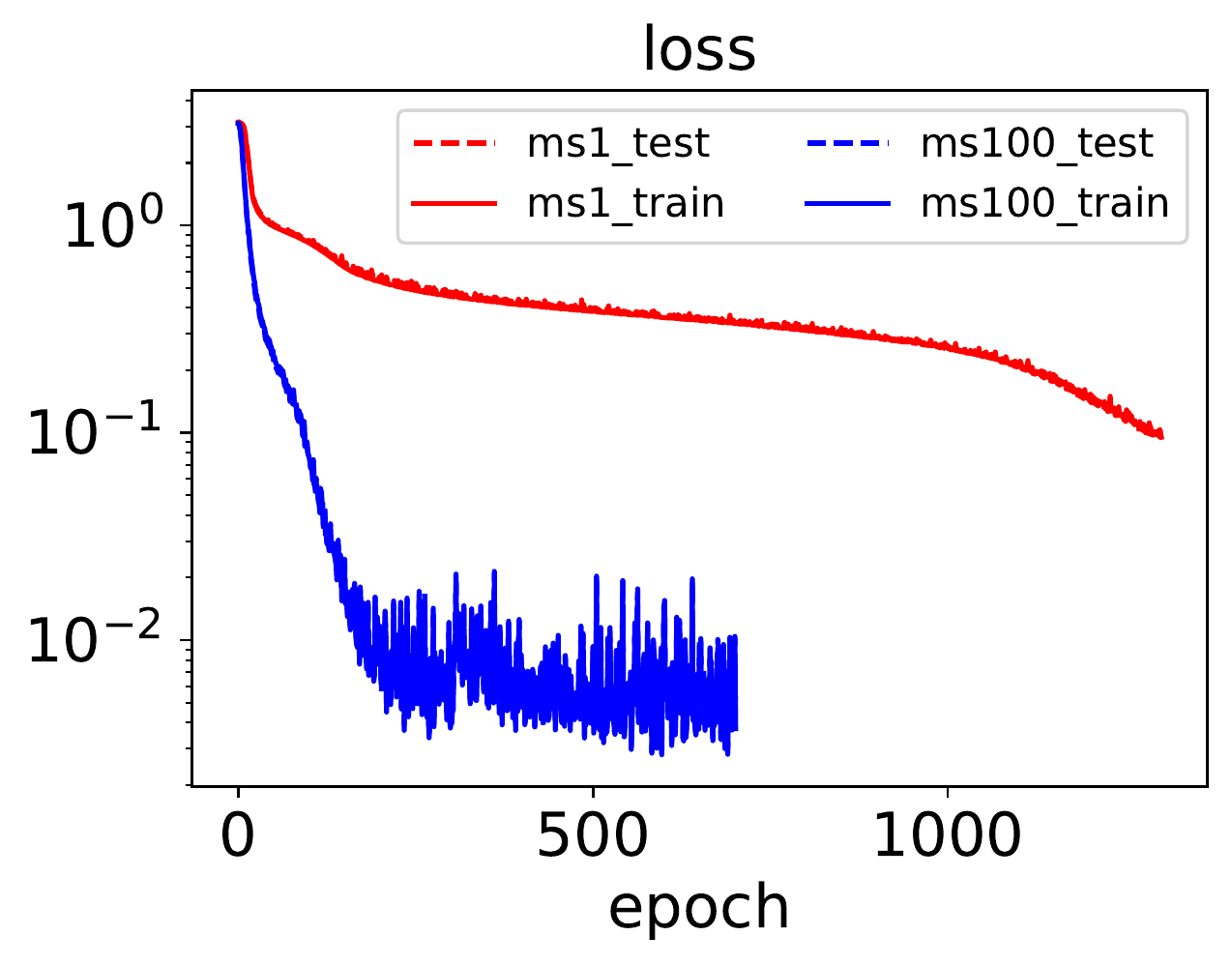}
	\par\end{centering}}
	\subfloat[$\ReLU(x)$, linear]{\begin{centering}
				\includegraphics[scale=0.35]{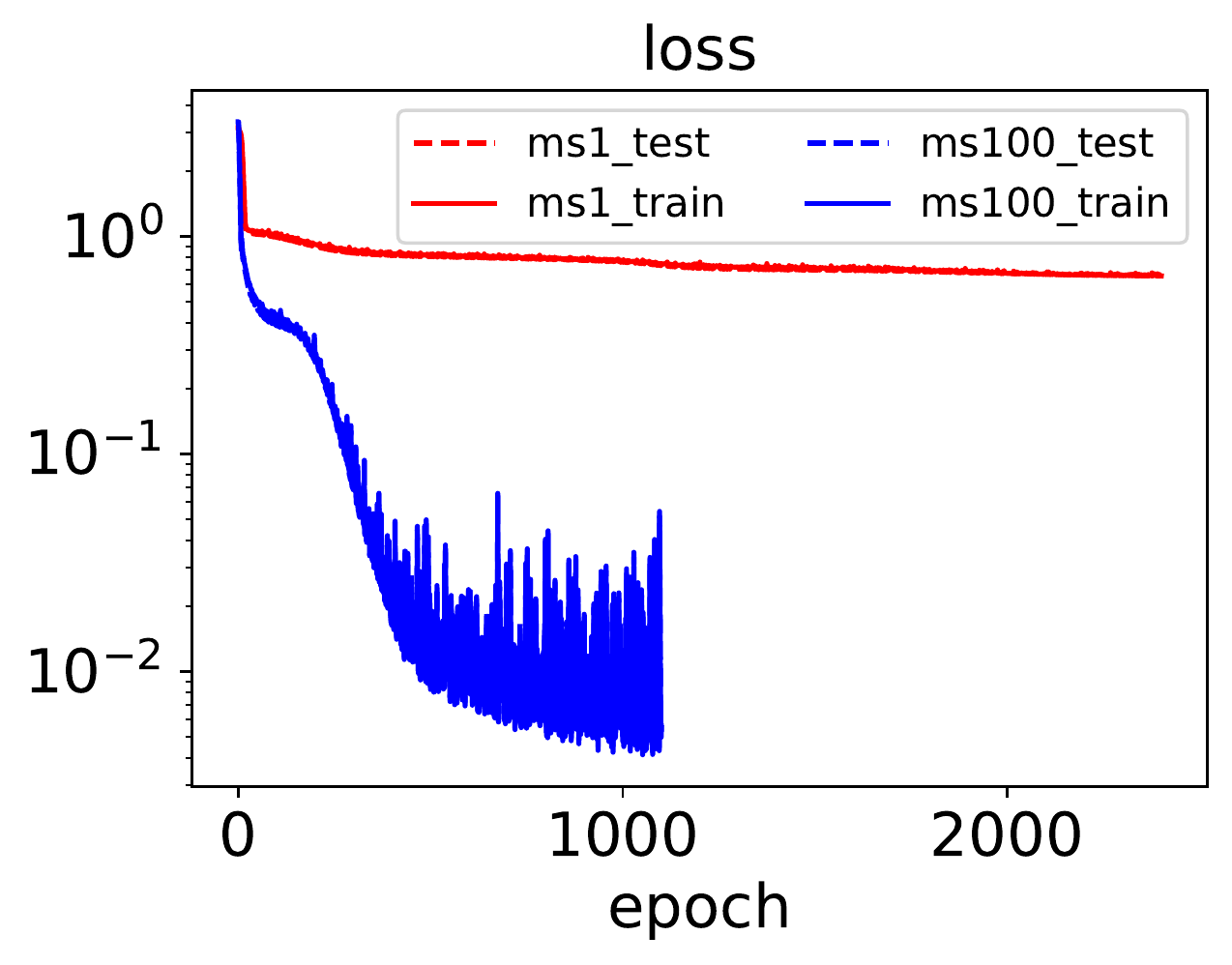}
	\par\end{centering}}

\par\end{centering}
\caption{Loss function vs. training epoch.   We use a network 60-200-200-200-1 with activation function  $\sReLU(x)$ or  $\ReLU(x)$. The learning rate is  $5\times10^{-5}$ with a decay rate  $2\times10^{-7}$ for each  training step with batch size 100. The training and test dataset are both $10000$ random samples.  Weights are initialized by  ${\cal D}_{1}$.  \label{fig:DNN-highd}}
\end{figure}
\par\end{center}

\begin{center}
\begin{figure}[h]
\begin{centering}
	\subfloat[$\sReLU(x)$, linear]{\begin{centering}
				\includegraphics[scale=0.35]{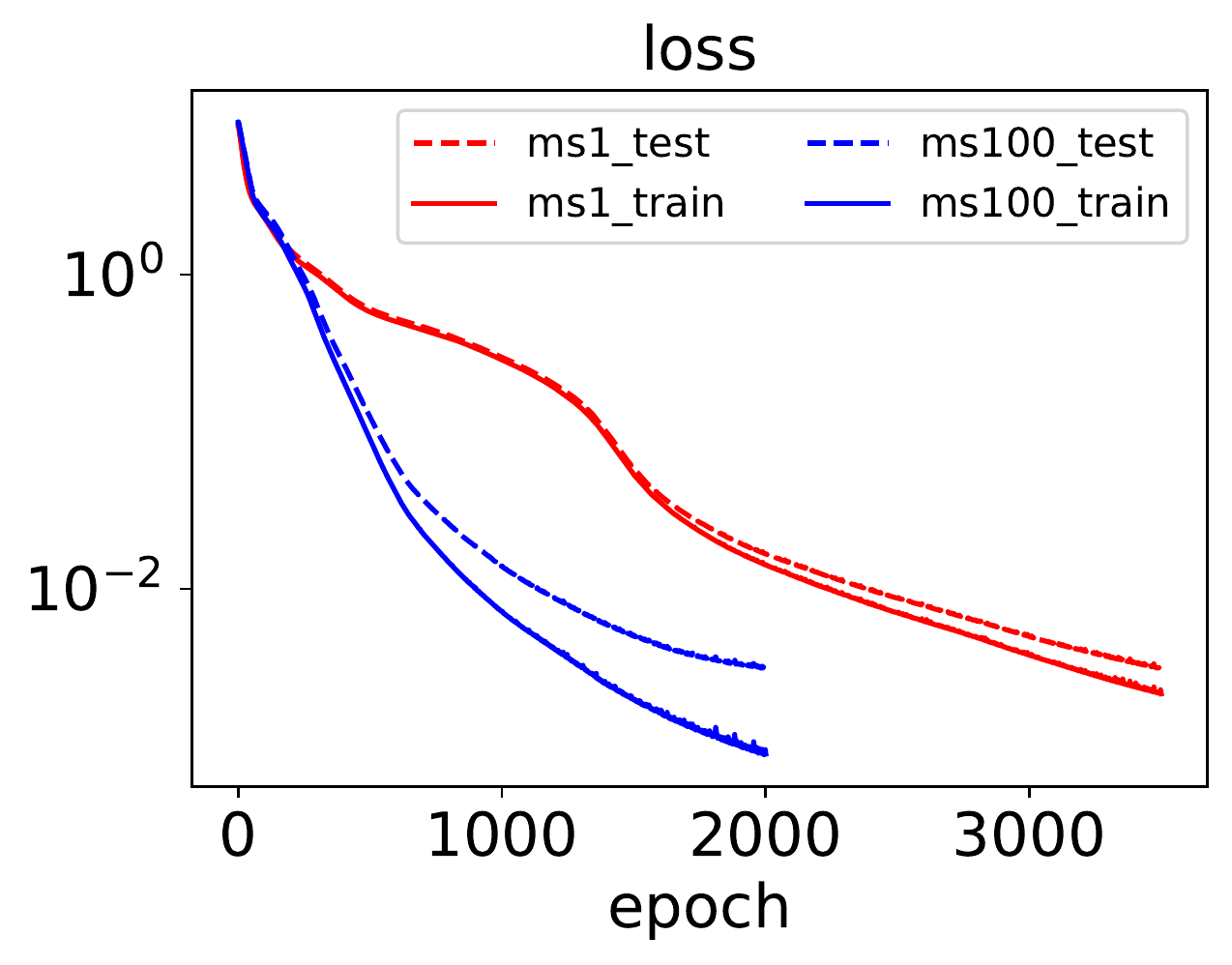}
	\par\end{centering}}
	\subfloat[$\sReLU(x)$, non-linear]{\begin{centering}
				\includegraphics[scale=0.35]{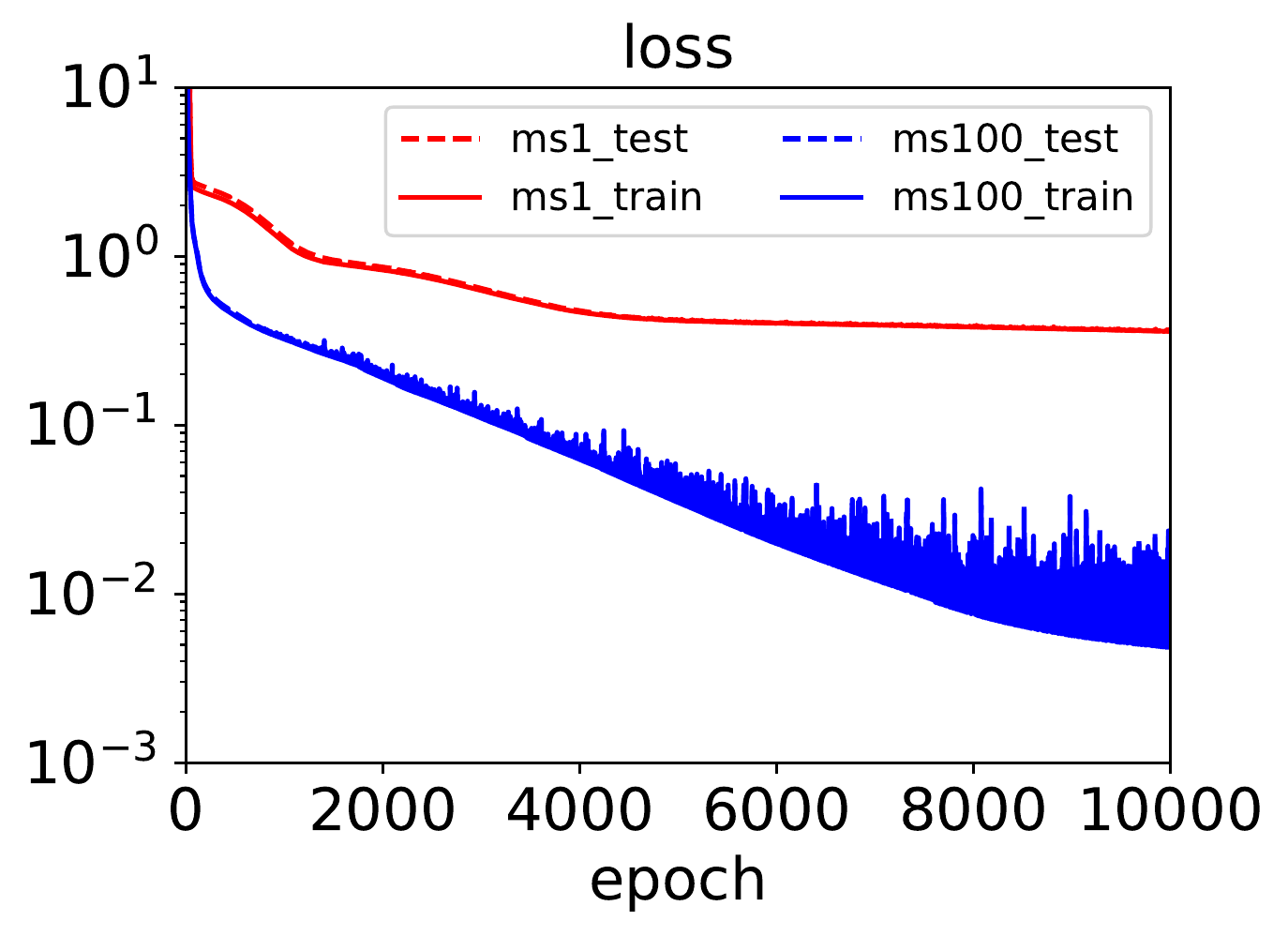}
	\par\end{centering}}
	\subfloat[$\ReLU(x)$, linear]{\begin{centering}
				\includegraphics[scale=0.35]{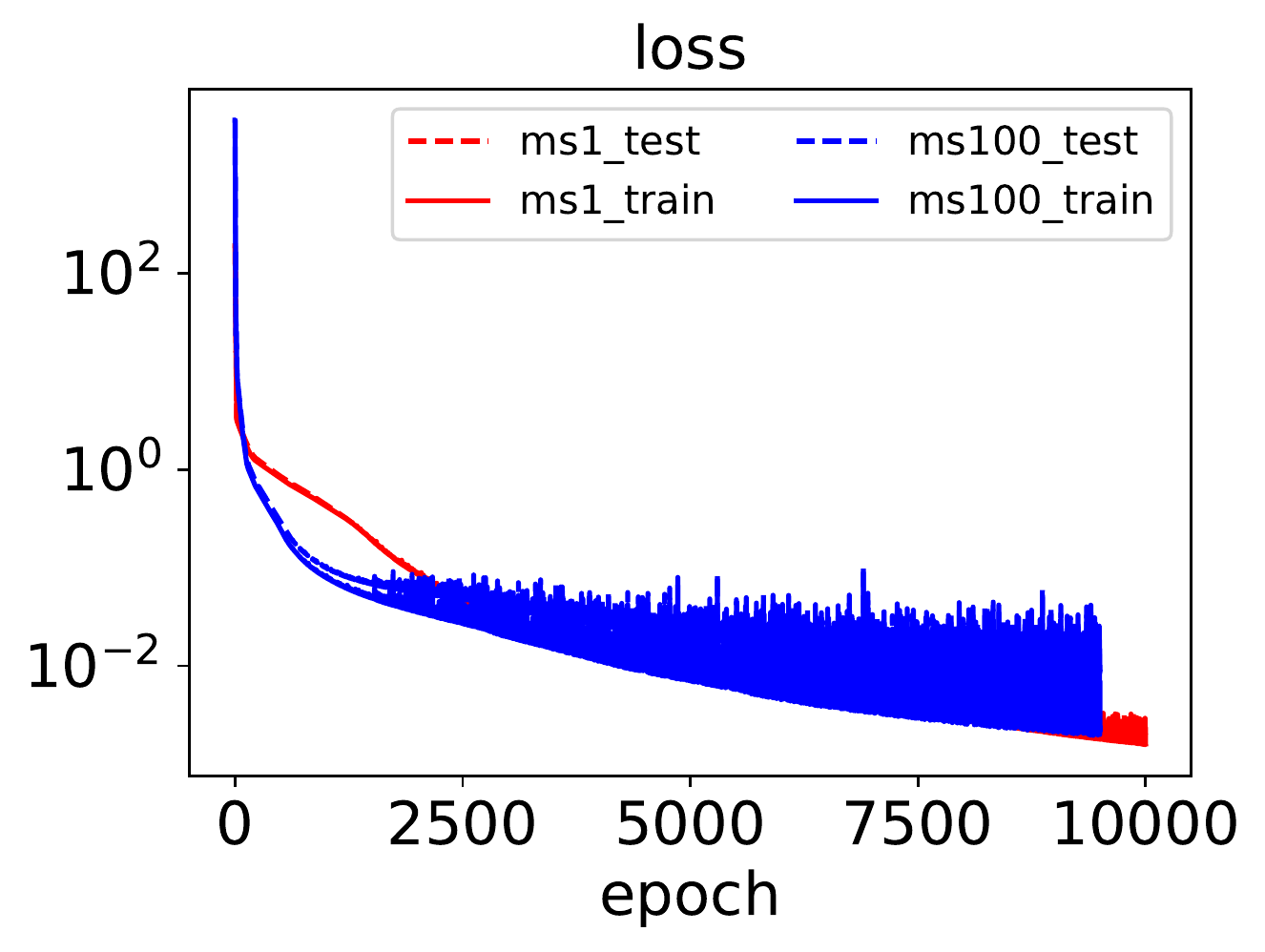}
	\par\end{centering}}

\par\end{centering}
\caption{Loss function vs. training epoch.   We use a network 60-500-500-500-500-1 with activation function  $\sReLU(x)$ or  $\ReLU(x)$. The learning rate is  $10^{-5}$ with a decay rate  $5\times10^{-7}$ for each  training step with batch size 1000. For the non-linear embedding example in (b), we use a smaller learning rate to avoid the fluctuation of loss function, $3\times 10^{-6}$.  The training and test dataset are both $5000$ random samples.  Weights are initialized by  ${\cal D}_{2}$.  \label{fig:DNN-highd2}}
\end{figure}
\par\end{center}

\subsection{Fitting high-frequency function}
Approximating high-frequency functions are very important in physical problems.  In this section, we use DNNs to fit high-frequency functions for 1d and 2d problems with MSE loss function. 
\paragraph{1d problems.} Consider 
 \[
 f(x)=\sin (23x)+\sin (137x)+\sin (203x), \quad x\in[0,\pi].
 \]
As shown in Fig.~\ref{fig:caiex1d}, with only one-scale DNN, the loss function decreases very slow during the training, while the loss function of the MscaleDNN decrease rapidly. We visualize the learned curves on test data points in Fig.~\ref{fig:caiex1dfunc}. For the one-scale DNN (first row in  Fig.~\ref{fig:caiex1dfunc}), the DNN learns a low-frequency function which cannot capture the target function. On the contrary, the MscaleDNN (second row in  Fig.~\ref{fig:caiex1dfunc}) accurately captures the highly oscillation of the target function.

\paragraph{2d problems.} Consider
\[
f(x,y)=f_{1}(x)f_{1}(y),
 \]
where 
\[
 f_{1}(x)=\sin (23x)+\sin (32x), \quad x\in[0,\pi].
\]
 As shown in Fig. \ref{fig:caiex2d}, similarly, the loss functions of MscaleDNN decays  much   faster than that of one-scale DNN during the training. On the test data points, Fig.~\ref{fig:caiex2dfunc} shows that the MscaleDNN predicts the target function well while one-scale DNN can not. To visualize the MscaleDNN can well predict the oscillation of the target function, we plot the true function and the DNN output at $y=0.5$ on test data points, as shown in Fig.~\ref{fig:caiex2dfunc}. The common one-scale DNN only captures the low-frequency oscillation, while the MscaleDNN captures well the highly oscillation of the target function. 
 
\begin{center}
	\begin{figure}
		\begin{centering}
			
			\includegraphics[scale=0.34]{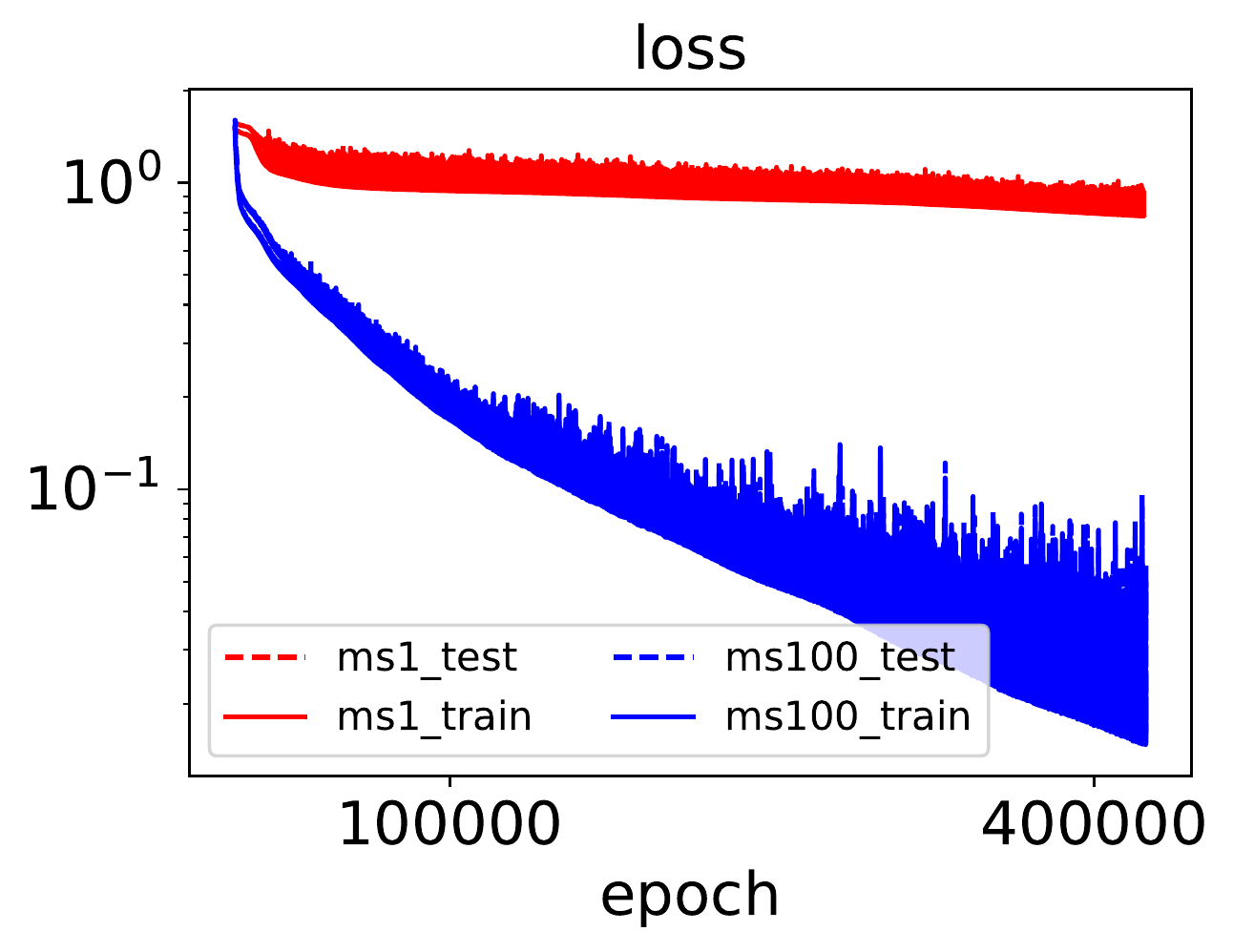} 
			
		\par\end{centering}
		\caption{ Loss function vs. training epoch.    Fitting  high-frequency 1d function.    We use a network 1-1000-500-500-500-500-1 with activation function  $\sReLU(x)$. The learning rate is  $10^{-5}$ with a decay rate  $5\times10^{-7}$ for each  training step with batch size 1000. The training and test dataset for 1d are   $5000$ and $1000$ random samples, respectively. Weights are initialized by  ${\cal D}_{2}$. \label{fig:caiex1d} }
	\end{figure}
\par\end{center}

\begin{center}
	\begin{figure}
		\begin{centering}
			
			\subfloat[full curve]{\begin{centering}
				\includegraphics[scale=0.34]{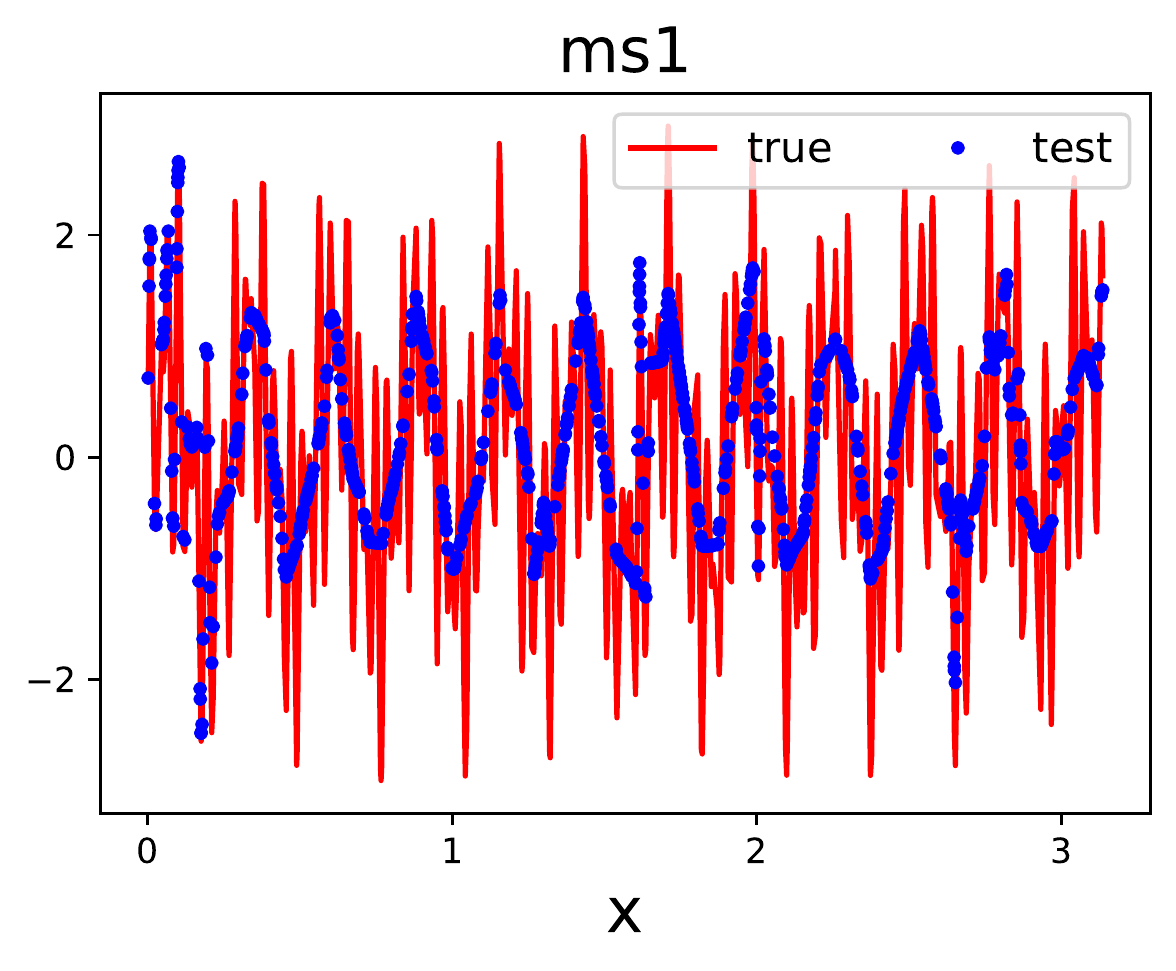} 
			\par\end{centering}}
			\subfloat[zoom in]{\begin{centering}
				\includegraphics[scale=0.34]{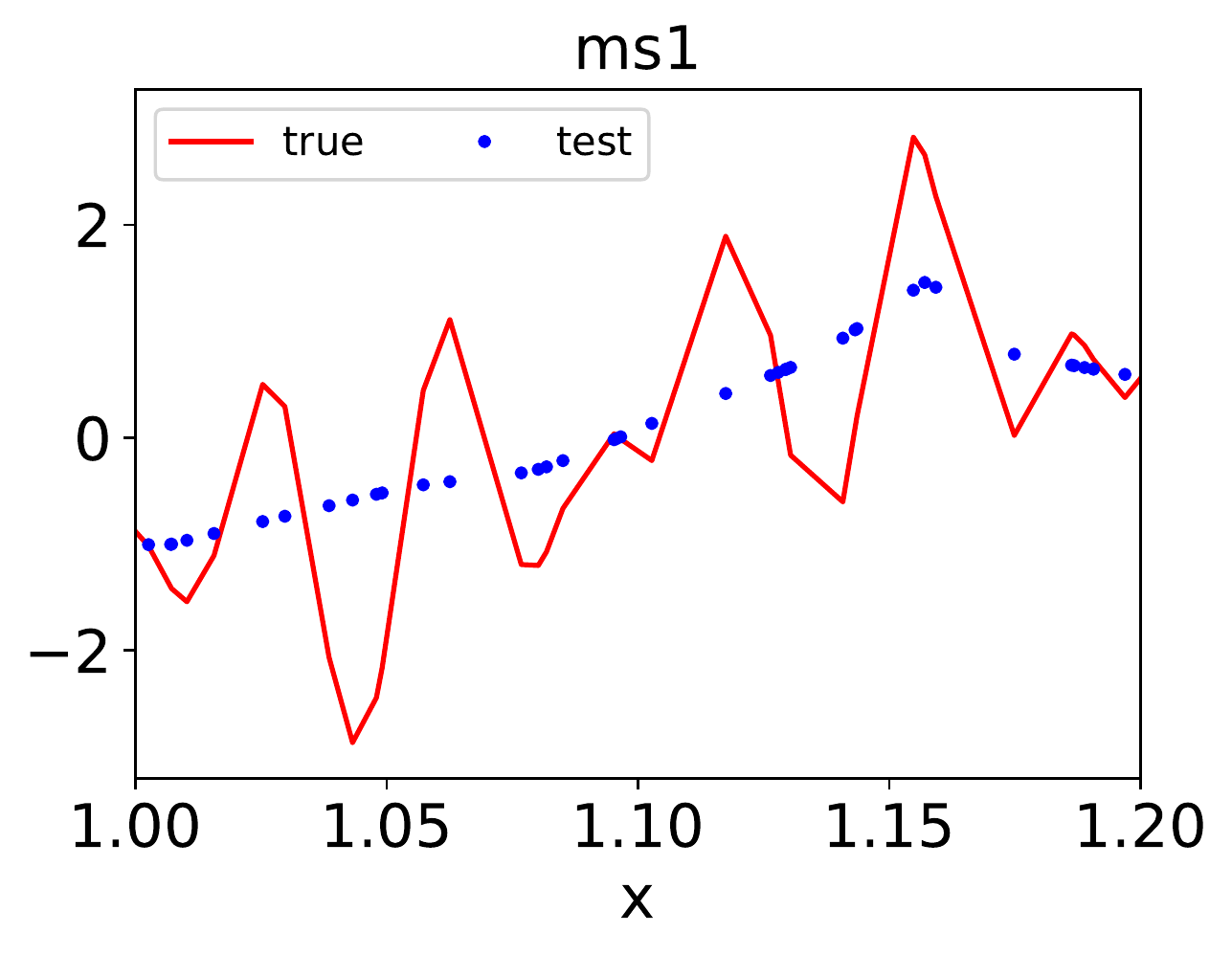} 
			\par\end{centering}}
			
			\subfloat[full curve]{\begin{centering}
				\includegraphics[scale=0.34]{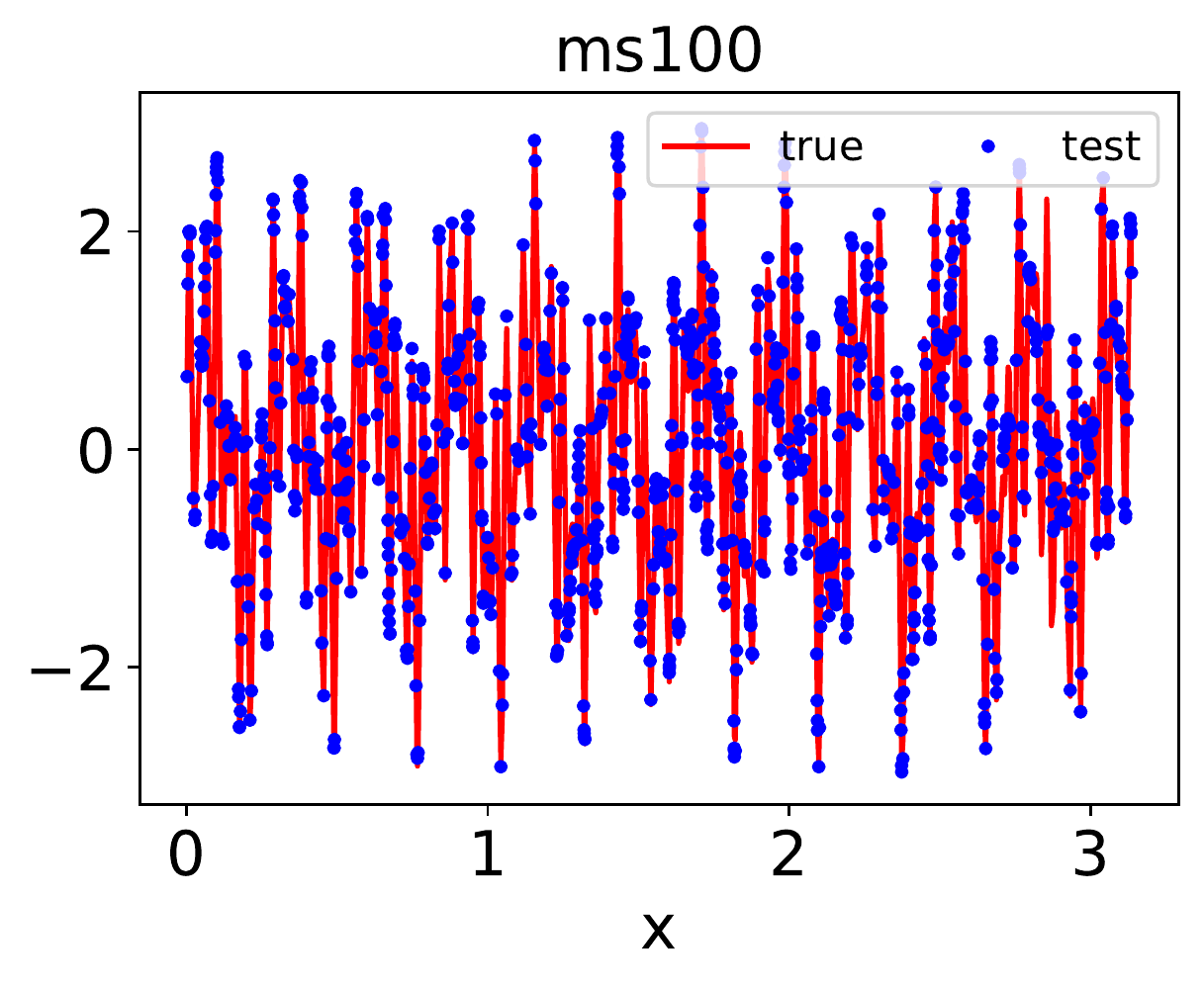} 
			\par\end{centering}}
			\subfloat[zoom in]{\begin{centering}
				\includegraphics[scale=0.34]{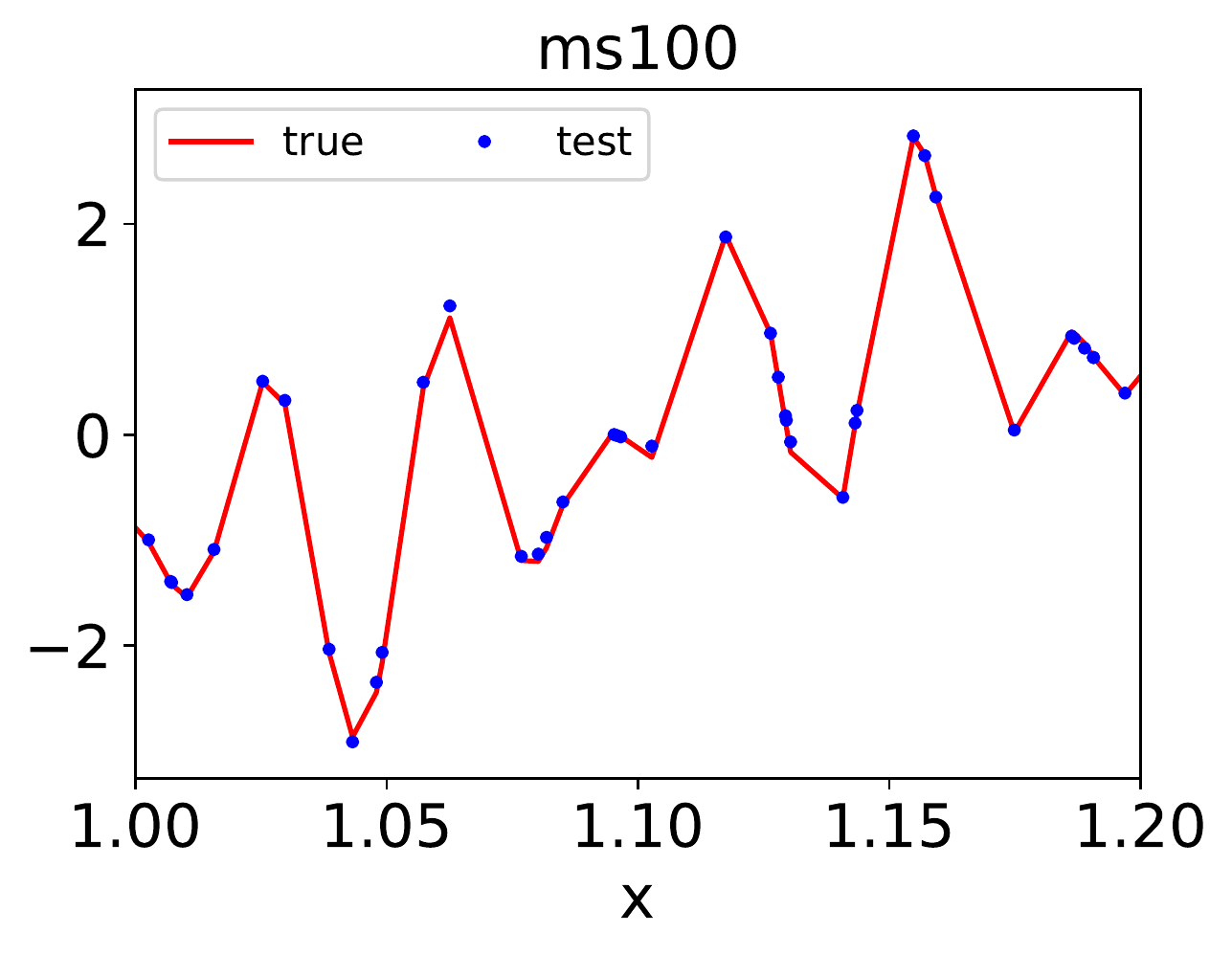} 
			\par\end{centering}}
			
		\par\end{centering}
		\caption{ The learning curves in fitting  high-frequency 1d  function on test data points.   \label{fig:caiex1dfunc} }
	\end{figure}
\par\end{center}

\begin{center}
	\begin{figure}
		\begin{centering}
			\includegraphics[scale=0.34]{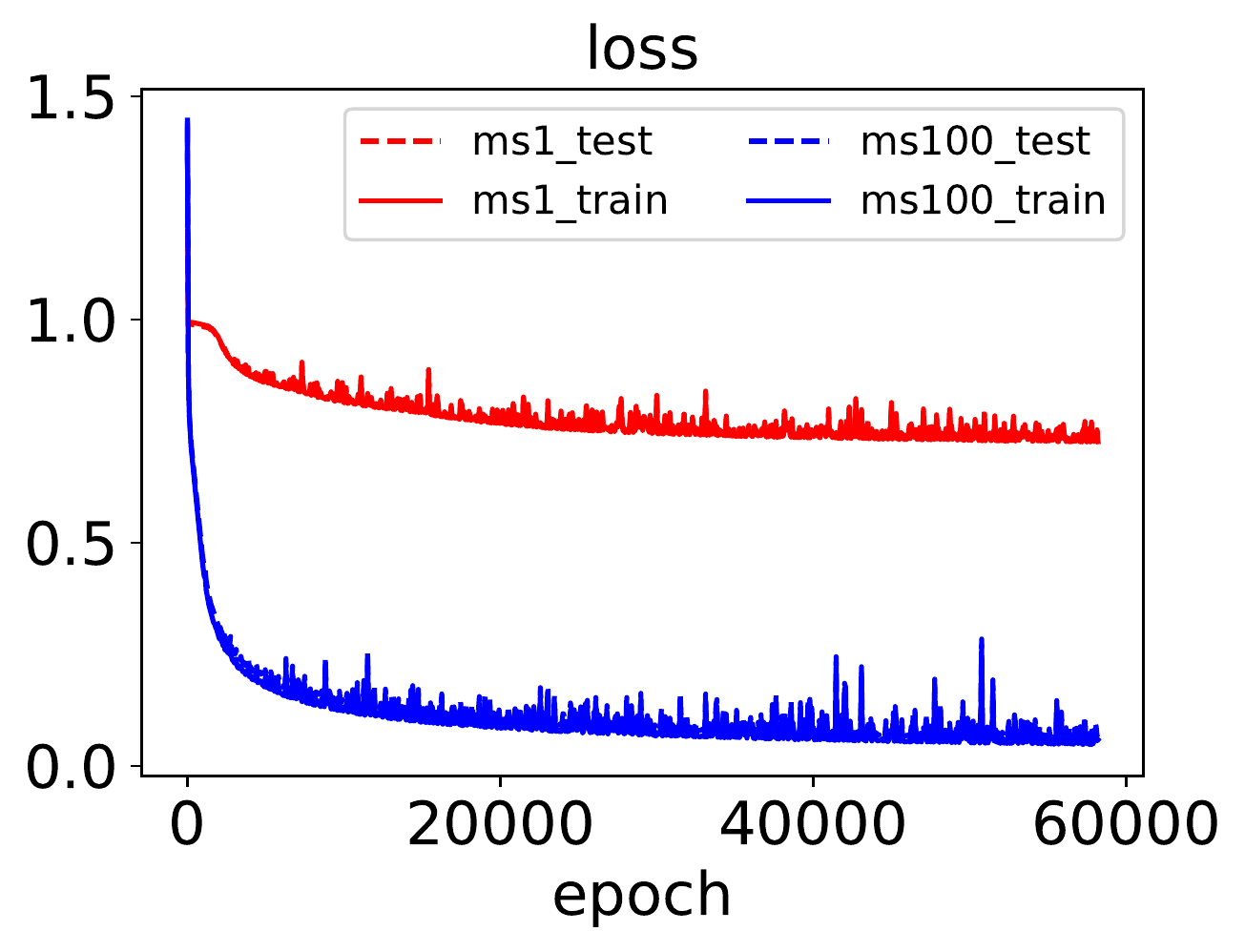} 
		\par\end{centering}
		\caption{Loss function vs. training epoch.    Fitting  high-frequency 2d function.    We use a network 2-1000-500-500-500-500-1 with activation function  $\sReLU(x)$. The learning rate is  $10^{-5}$ and  $2\times 10^{-5}$ for 1d and 2d, respectively, with a decay rate  $5\times10^{-7}$ for each  training step with batch size 10000. The training and test dataset  are  $100000$ and $50000$ random samples, respectively. Weights are initialized by  ${\cal D}_{2}$.\label{fig:caiex2d} }
	\end{figure}
\par\end{center}

\begin{center}
	\begin{figure}
		\begin{centering}
			
			\subfloat[ms1]{\begin{centering}
				\includegraphics[scale=0.34]{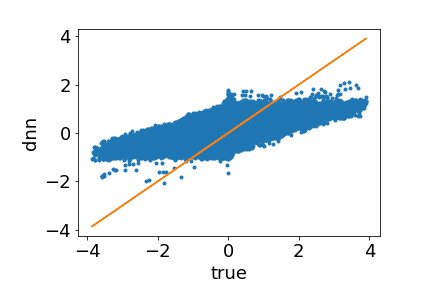} 
			\par\end{centering}}
			\subfloat[ms100]{\begin{centering}
				\includegraphics[scale=0.34]{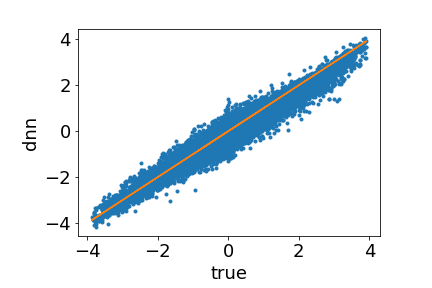} 
			\par\end{centering}}
			
			\subfloat[ms1]{\begin{centering}
				\includegraphics[scale=0.34]{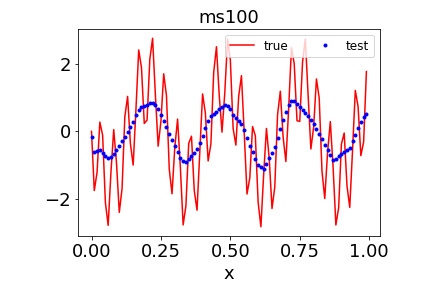} 
			\par\end{centering}}
			\subfloat[ms100]{\begin{centering}
				\includegraphics[scale=0.34]{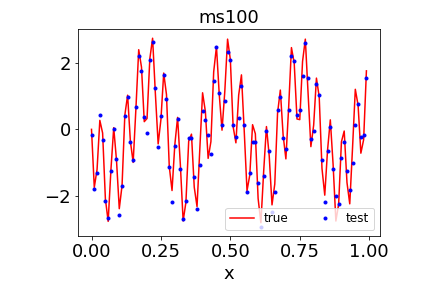} 
			\par\end{centering}}
			
		\par\end{centering}
		\caption{ The learning curves in fitting  high-frequency 2d  function. First row:  Each point corresponds to a test data point, the abscissa is the true value and the ordinate is the DNN output. Second row: The line of $y=0.5$ for the true function and the DNN output at test data points. \label{fig:caiex2dfunc} }
	\end{figure}
\par\end{center}

\subsection{Solving high dimensional PDE}

An important application of MscaleDNN is to solve high-dimensional PDEs, which is often very difficult for traditional methods due to the curse of dimensionality. A previous work \citep{xu2019frequency} proposes to use DNNs to capture the low-frequency part of the solution and then use the DNN output after training as the initial value for traditional methods to continuously learn the high-frequency part. Later, another work \citep{huang2019int} shows that by using DNN output as initialization for traditional methods can accelerate the solving process of many PDEs. The PhaseDNN attempts to overcome the difficulty of learning high frequency by shifting high frequency to low frequency in each coordinate.  However, all these methods suffer from the curse of dimensionality. The Deep Ritz method proposed to solve PDEs with DNNs  in \cite{weinan2017deep} may overcome the curse of dimensionality, however, they are not able to handle high-frequency functions efficiently due to the F-Principle bahavior of traditional DNNs. The MscaleDNN uses multiple scales to facilitate the learning of many frequencies while having the potential to overcome the curse of dimensionality. 

We consider the following example, 
\begin{equation}
g(\vx)=\sum_{i=1}^{d}\sin(x_{i})+100\sin(10x_{i})
\end{equation}
and 
\begin{equation}
\tilde{g}(\vx)=\sum_{i=1}^{d}\sin(x_{i})+\sin(10x_{i}), \quad \vx \in \partial\Omega
\end{equation}

The true solution is 
\begin{equation}
u_{\rm true}(\vx)=\sum_{i=1}^{d}\sin(x_{i})+\sin(10x_{i}).
\end{equation}

Results with both Ritz loss in Eq.~(\ref{ritzlossnum}) and LSE loss in Eq.~(\ref{lselossnum}) are given below.

\paragraph{Ritz loss}

We show  ${\rm MSE}(h(\vx),u_{\rm true}(\vx))$ during the training process for $d=3, 10, 25$ in Fig.~\ref{fig:ritzsgd} with initialization ${\cal D}_{1}$ and in Fig.~\ref{fig:ritzsgd2} with initialization ${\cal D}_{2}$. In all cases, MscaleDNNs with $100$ scales are much faster than that of only one scale. Note that if we only use the activation function of $\sigma(100\vw\cdot\vx+b)$ for the first hidden layer, the learning is very slow. 

\begin{center}
	\begin{figure}
		\begin{centering}
			
			\subfloat[3d]{\begin{centering}
				\includegraphics[scale=0.34]{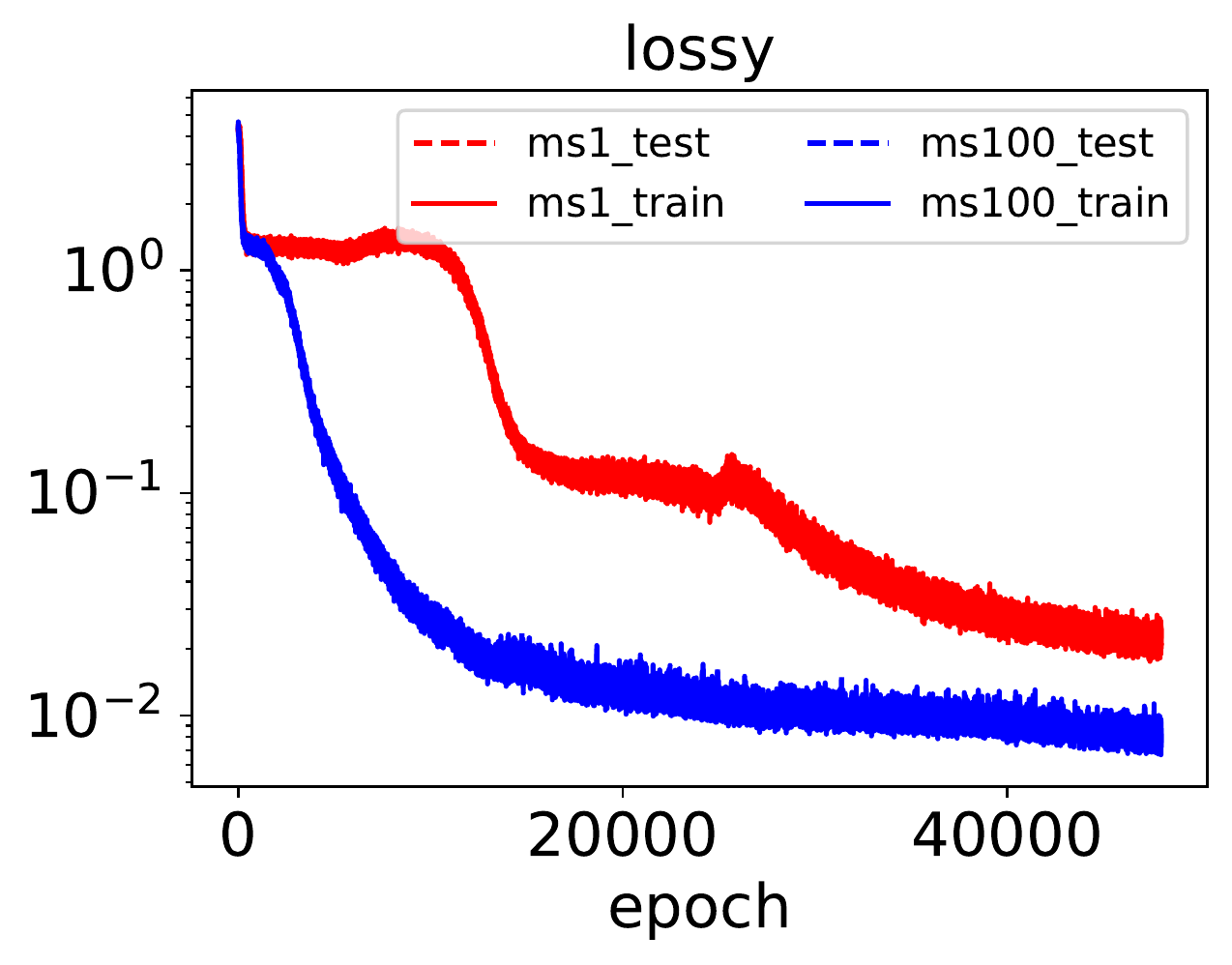} 
			\par\end{centering}}
			\subfloat[10d]{\begin{centering}
				\includegraphics[scale=0.34]{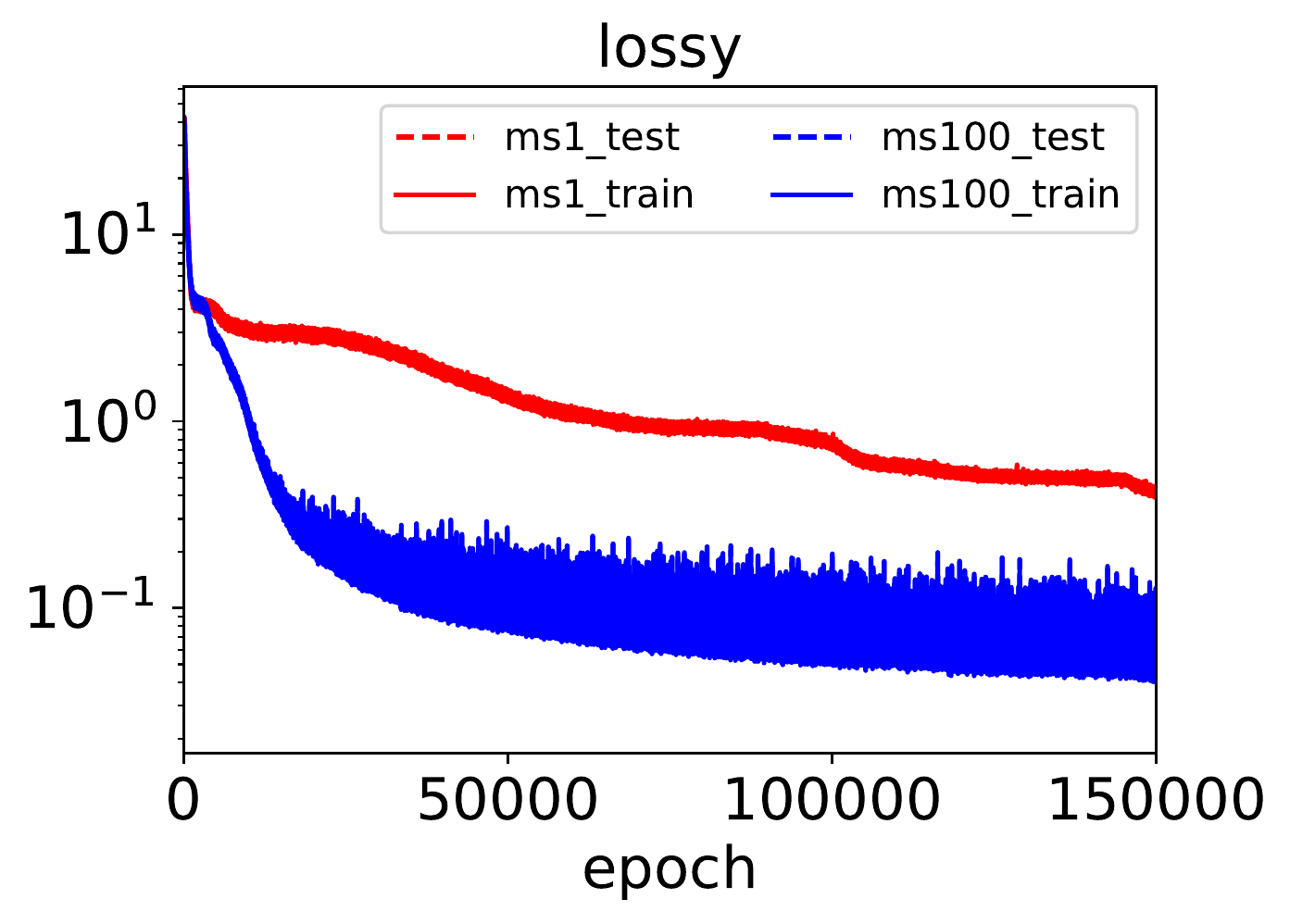} 
			\par\end{centering}}
			\subfloat[25d]{\begin{centering}
				\includegraphics[scale=0.34]{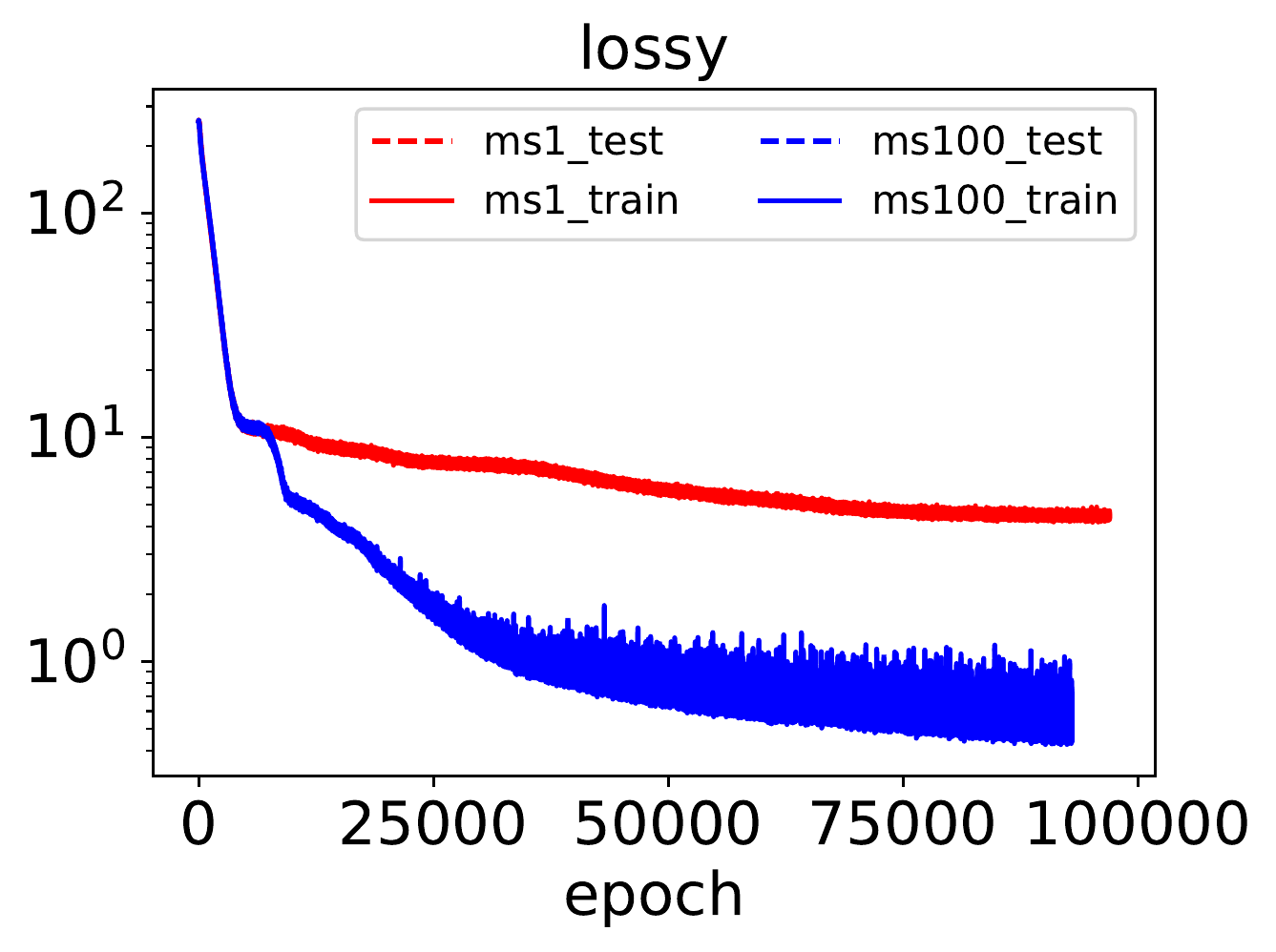} 
			\par\end{centering}}
			
		\par\end{centering}
		\caption{Ritz loss: ${\rm MSE}(h(\vx),u_{\rm true}(\vx))$ vs. training epoch.  ${\rm sReLU}(x)$. Ritz loss. $\Omega=[0,1]^d$, $n=1000$, $\tilde{n}=100$,  $\beta=1000$, we use a network d-200-200-200-1 with activation function $\sReLU(x)$. The learning rate is  $5\times10^{-5}$ with a decay rate  $5\times10^{-7}$ for each training step.  Weights are initialized by  ${\cal D}_{1}$. \label{fig:ritzsgd} }
	\end{figure}
\par\end{center}

\begin{center}
	\begin{figure}
		\begin{centering}
			
			\subfloat[3d]{\begin{centering}
				\includegraphics[scale=0.34]{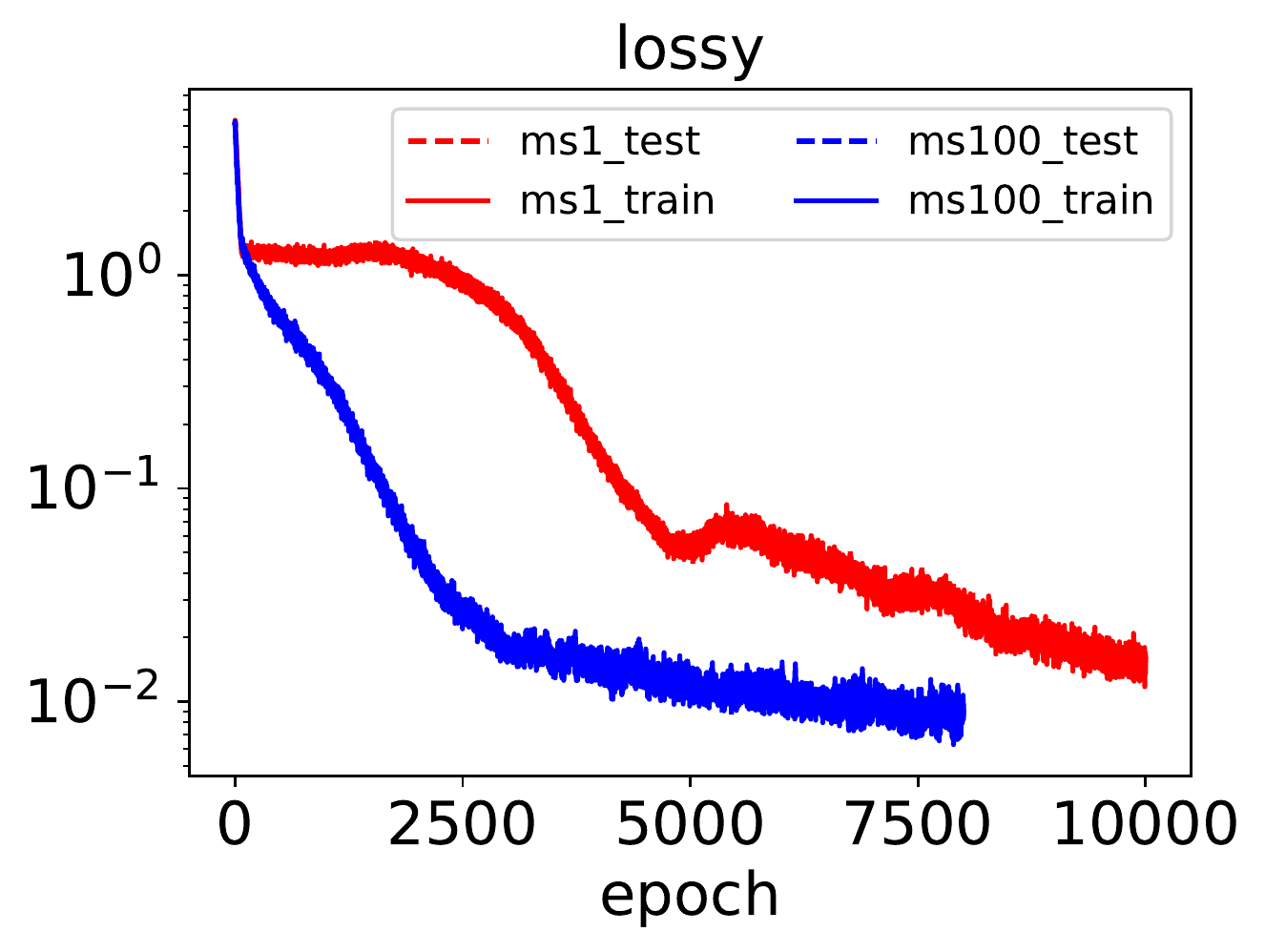} 
			\par\end{centering}}
			\subfloat[10d]{\begin{centering}
				\includegraphics[scale=0.34]{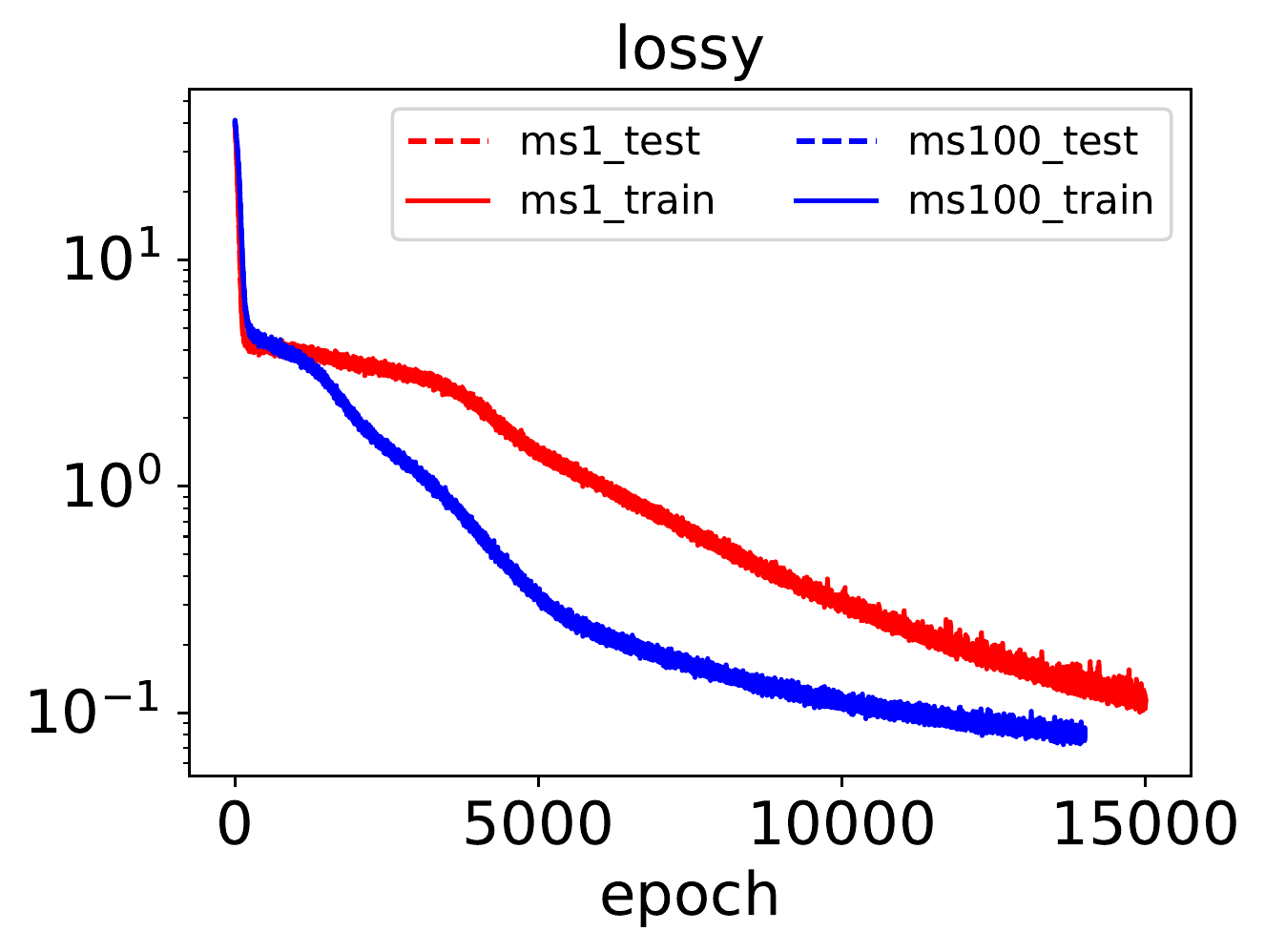} 
			\par\end{centering}}
			\subfloat[25d]{\begin{centering}
				\includegraphics[scale=0.34]{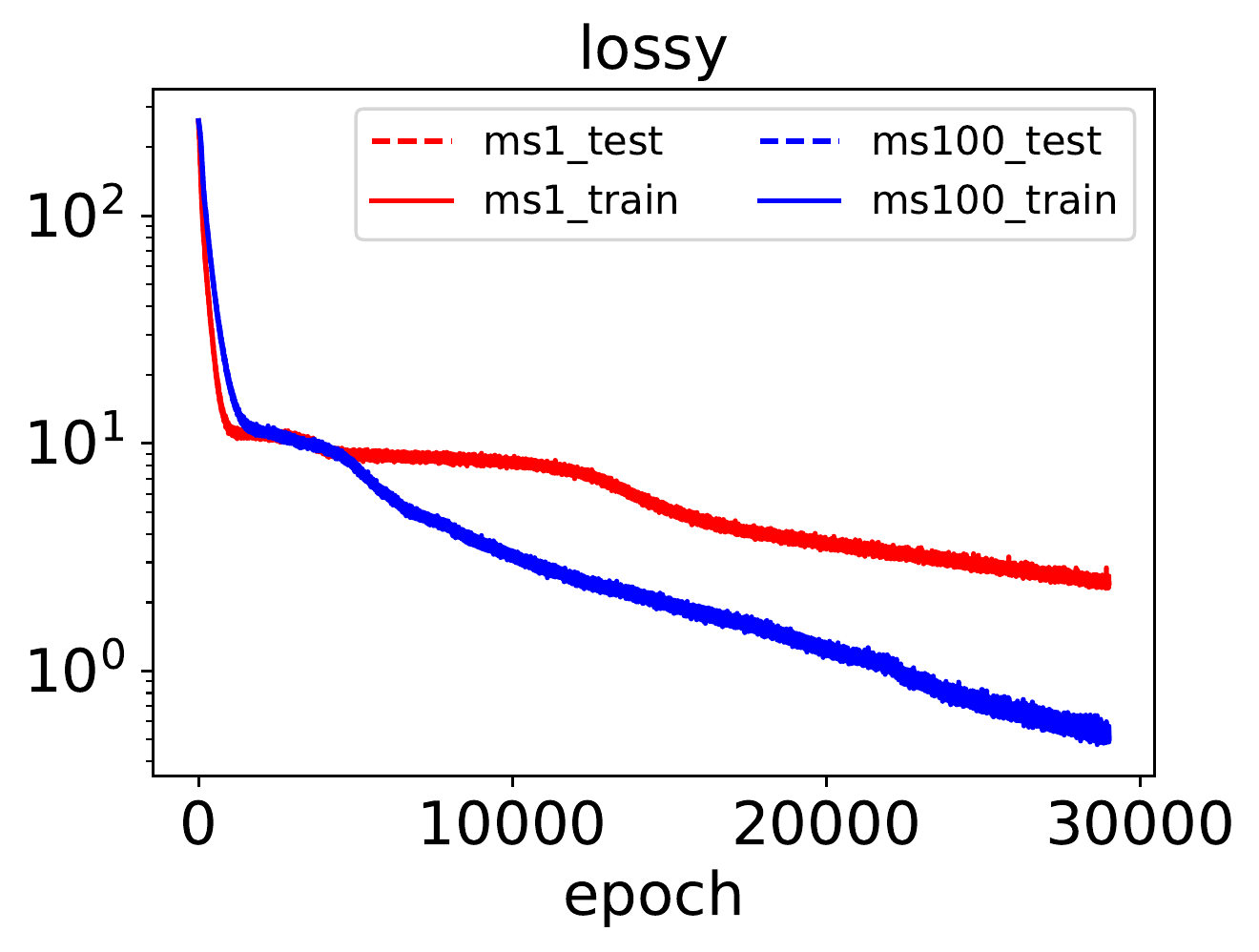} 
			\par\end{centering}}
			
		\par\end{centering}
		\caption{Ritz loss: ${\rm MSE}(h(\vx),u_{\rm true}(\vx))$ vs. training epoch.  ${\rm sReLU}(x)$. Ritz loss. $\Omega=[0,1]^d$, $n=1000$, $\tilde{n}=100$, $\beta=1000$, we use a network d-500-500-500-1 with activation function $\sReLU(x)$. The learning rate is  $5\times10^{-5}$ with a decay rate  $5\times10^{-7}$ for each training step.  Weights are initialized by  ${\cal D}_{2}$. \label{fig:ritzsgd2} }
	\end{figure}
\par\end{center}

\paragraph{LSE loss}

We show  ${\rm MSE}(h(\vx),u_{\rm true}(\vx))$ during the training process for $d=3, 10$ in Fig.~\ref{fig:poindgd} with initialization ${\cal D}_{1}$ and in Fig.~\ref{fig:poindgd2} with initialization ${\cal D}_{2}$. In all cases, MscaleDNNs with $100$ scales are much faster than that of only one scale.   Note that since the LSE loss requires Laplacian operation, its computational cost is much higher.

\begin{center}
	\begin{figure}
		\begin{centering}
			
			\subfloat[3d]{\begin{centering}
				\includegraphics[scale=0.34]{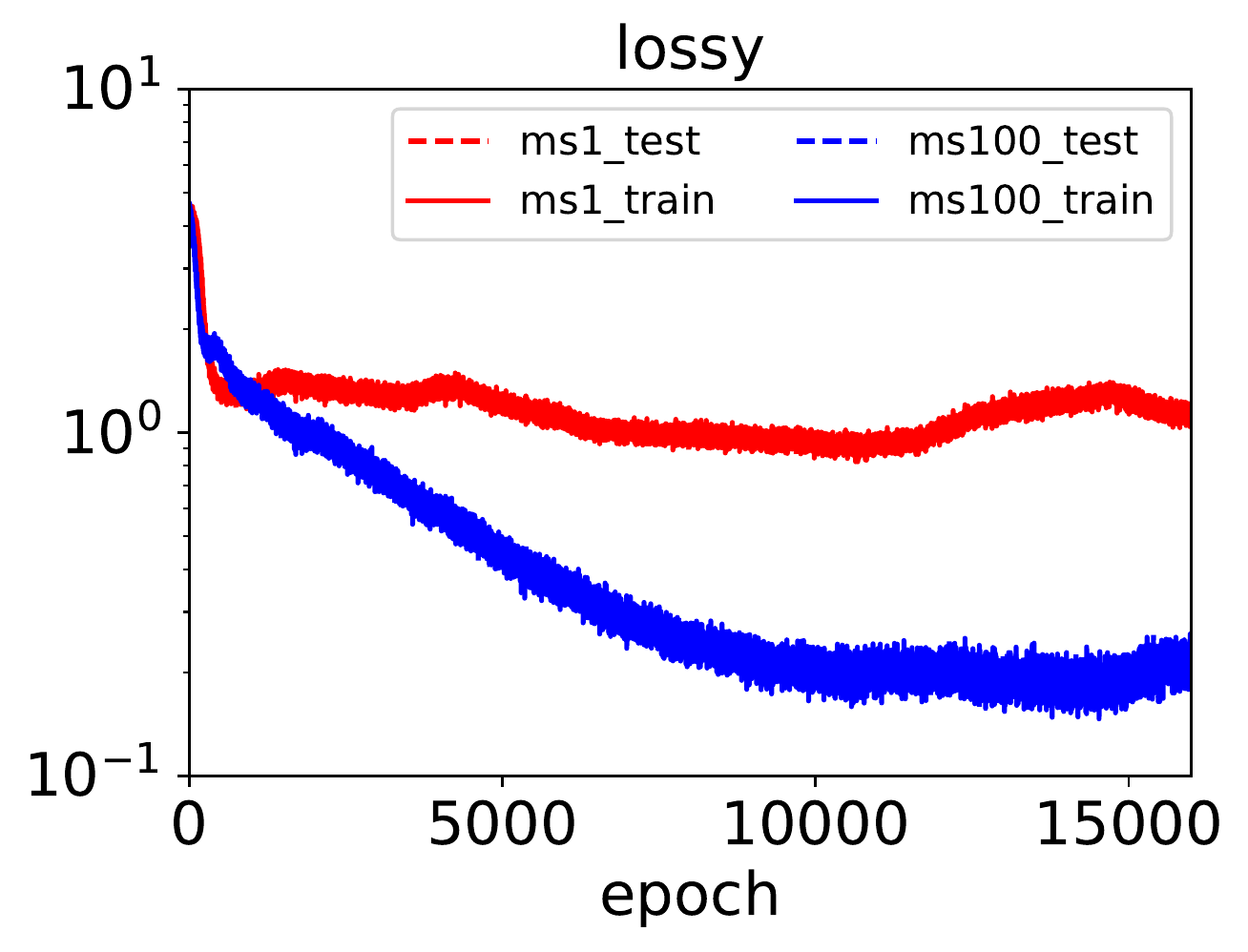} 
			\par\end{centering}}
			\subfloat[10d]{\begin{centering}
				\includegraphics[scale=0.34]{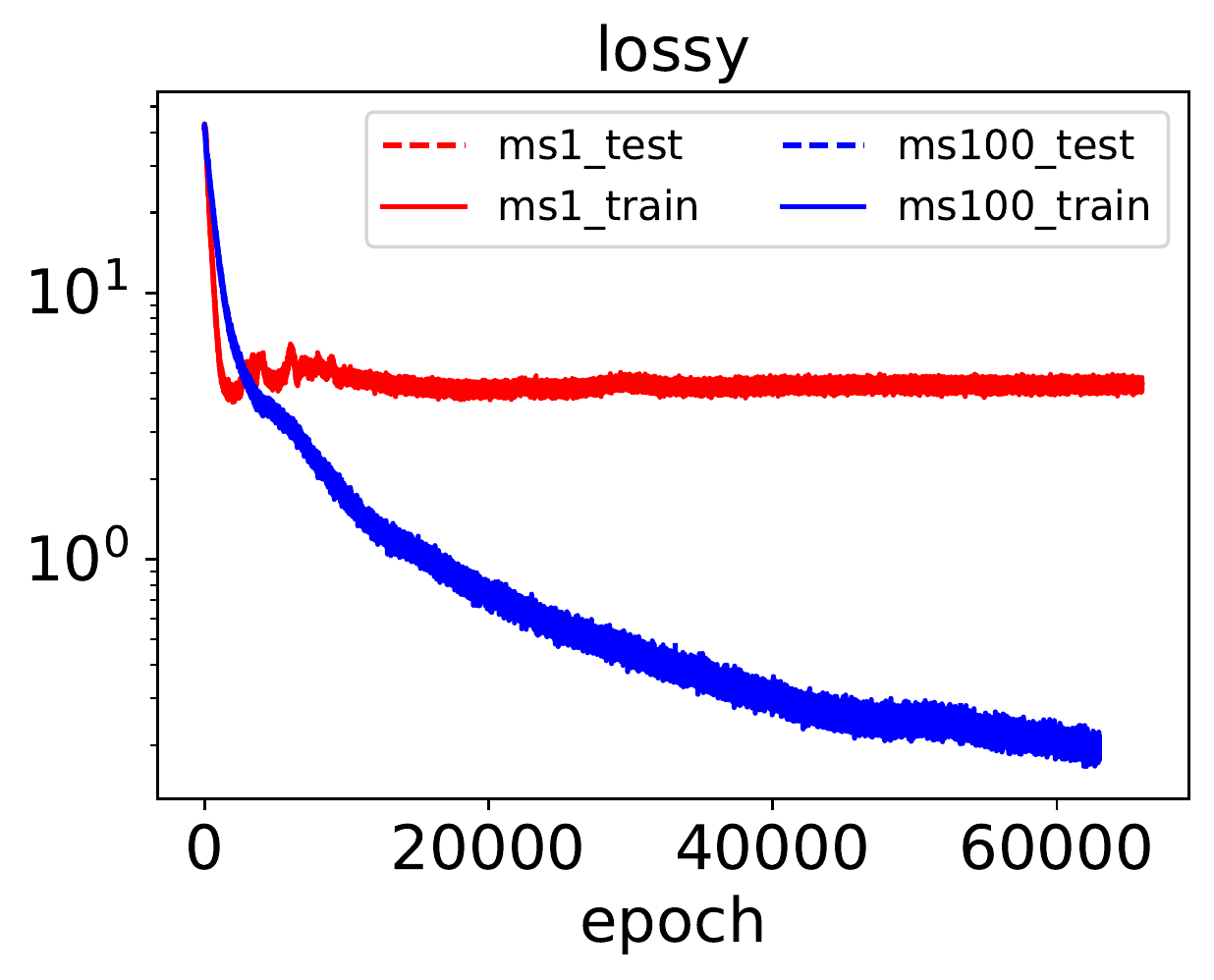} 
			\par\end{centering}}
			
		\par\end{centering}
		\caption{LSE loss: ${\rm MSE}(h(\vx),u_{\rm true}(\vx))$ vs. training epoch.   Direct loss. $\Omega=[0,1]^d$, $n=1000$, $\tilde{n}=100$,  $\beta=1000$, we use a network d-200-200-200-1 with activation function $\sReLU(x)$. The learning rate is  $5\times10^{-5}$ with a decay rate  $5\times10^{-7}$ for each training step.   Weights are initialized by  ${\cal D}_{1}$.   \label{fig:poindgd} }
	\end{figure}
\par\end{center}

\begin{center}
	\begin{figure}
		\begin{centering}
			
			\subfloat[3d]{\begin{centering}
				\includegraphics[scale=0.34]{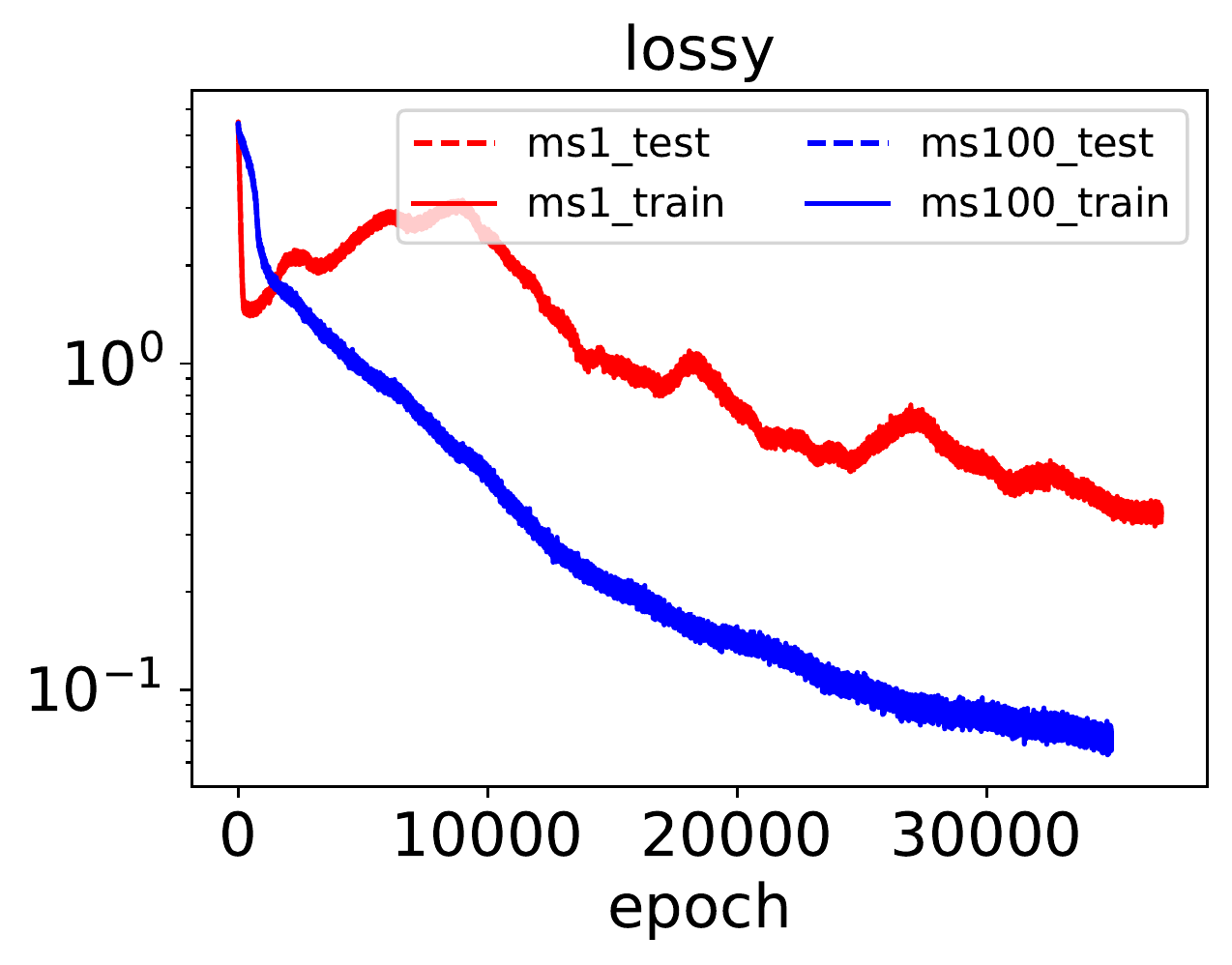} 
			\par\end{centering}}
			\subfloat[10d]{\begin{centering}
				\includegraphics[scale=0.34]{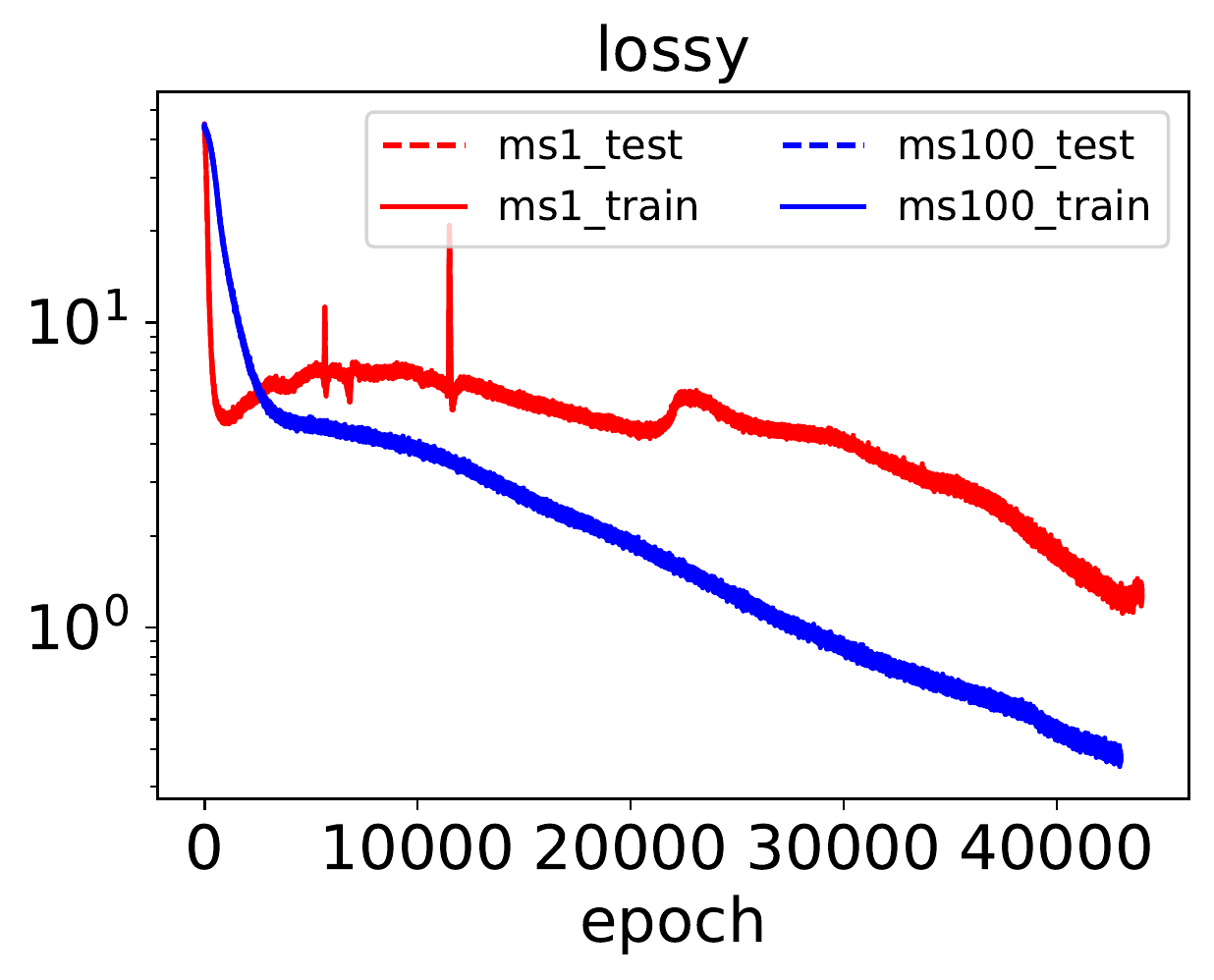} 
			\par\end{centering}}
			
		\par\end{centering}
		\caption{LSE loss: ${\rm MSE}(h(\vx),u_{\rm true}(\vx))$ vs. training epoch.   Direct loss.  $\Omega=[0,1]^d$, $n=5000$, $\tilde{n}=100$,  $\beta=1000$, we use a network d-200-200-200-1 with activation function $\sReLU(x)$. The learning rate is  $5\times10^{-5}$ with a decay rate  $5\times10^{-7}$ for each training step.   Weights are initialized by  ${\cal D}_{2}$.   \label{fig:poindgd2} }
	\end{figure}
\par\end{center}

\section{Conclusion and future work}
In this paper, we have introduced a frequency domain scaling DNN with compact supported activation function,  MscaleDNN, to produce a multi-scale DNN for approximating function of high frequency and high dimensions as well as the solution of PDEs in high dimensions. By using the radial scaling in the $k$-space of the functions, we are able to achieve much faster learning results for high dimensional function fitting and solution of elliptic problems though either least square residual or Ritz energy minimization. While the Ritz minimization approach only requires the first derivatives of the DNN solution, the least square residual, which does need to compute the full Laplacian of the DNN, thus incurring a higher cost, can be applicable to non-self adjoint differential operators, such as high dimensional Fokker-Planck equations.

In the current implementation of MscaleDNN, we only used neurons with scaled compact supported activation functions, this corresponds to the scaling spaces in the wavelet theory. Because many scales are used in the first hidden layer of the MscaleDNN, we actually created much redundancy or overlapping of scales in the neurons. A more sophisticated way, following the idea of mother wavelet function \citep{daubechies1992ten}, is to introduce another type of neuron made of activation function with similar properties of the mother wavelet, such as vanishing moments, leading to a Wavelet-DNN multi-resolution framework for high dimensional problem.

\section{Acknowledgments}
The second author is supported by the Student Innovation Center at Shanghai Jiao Tong University and thanks Douglas Zhou (SJTU) for providing computational resource.

%\newpage{}

\bibliographystyle{agsm}
\bibliography{DLRef}

\newpage
\end{document}